\documentclass[10pt,twocolumn,letterpaper]{article}

\usepackage{cvpr}              

\usepackage{graphicx}
\usepackage{amsmath}
\usepackage{amssymb}
\usepackage{booktabs}
\usepackage{anyfontsize}
\usepackage{multirow}
\usepackage{subfig}

\usepackage{graphbox}
\usepackage{paralist}

\usepackage{makecell}
\usepackage{tabularx}

\usepackage{algorithm}
\usepackage{setspace}
\usepackage[noend]{algpseudocode}

\usepackage{color}
\usepackage[dvipsnames,svgnames]{xcolor}

\usepackage[normalem]{ulem}

\usepackage{float}

\usepackage[normalem]{ulem}

\definecolor{MyGreen}{cmyk}{100, 0, 100, 0}
\usepackage[pagebackref,breaklinks,colorlinks,citecolor=MyGreen]{hyperref}
\usepackage{pifont}
\usepackage{helvet}
\usepackage{caption}

\usepackage[capitalize]{cleveref}
\crefname{section}{Sec.}{Secs.}
\Crefname{section}{Section}{Sections}
\Crefname{table}{Table}{Tables}
\crefname{table}{Tab.}{Tabs.}

\newcommand{\fix}[1]{\textcolor{black}{#1}}

\newcommand{\figcspace}{\vspace{3mm}}
\newcommand{\figspace}{\vspace{-5mm}}
\newcommand{\tabcspace}{\vspace{3mm}}
\newcommand{\tabspace}{\vspace{-5mm}}
\usepackage{lipsum}

\newcommand\blfootnote[1]{%
  \begingroup
  \renewcommand\thefootnote{}\footnote{#1}%
  \addtocounter{footnote}{-1}%
  \endgroup
}

\usepackage[capitalize]{cleveref}
\crefname{section}{Sec.}{Secs.}
\Crefname{section}{Section}{Sections}
\Crefname{table}{Table}{Tables}
\crefname{table}{Tab.}{Tabs.}

\def\cvprPaperID{4764} 
\def\confName{CVPR}
\def\confYear{2022}

\begin{document}
\title{HandOccNet: Occlusion-Robust 3D Hand Mesh Estimation Network}

\author{JoonKyu Park$^{1*}$ \quad Yeonguk Oh$^{1*}$ \quad Gyeongsik Moon$^{1*}$ \quad Hongsuk Choi$^{1}$ \quad Kyoung Mu Lee$^{1,2}$ 
\\$^{1}$Dept. of ECE \& ASRI, $^{2}$IPAI, Seoul National University, Korea\\
{\tt\small jkpark0825@snu.ac.kr,namepllet1@gmail.com,\{mks0601,redarknight,kyoungmu\}@snu.ac.kr}}

\maketitle

\begin{abstract}
Hands are often severely occluded by objects, which makes 3D hand mesh estimation challenging.
Previous works often have disregarded information at occluded regions.
However, we argue that occluded regions have strong correlations with hands so that they can provide highly beneficial information for complete 3D hand mesh estimation.
Thus, in this work, we propose a novel 3D hand mesh estimation network HandOccNet, that can fully exploits the information at occluded regions as a secondary means to enhance image features and make it much richer.
To this end, we design two successive Transformer-based modules, called feature injecting transformer (FIT) and self-enhancing transformer (SET).
FIT injects hand information into occluded region by considering their correlation.
SET refines the output of FIT by using a self-attention mechanism.
By injecting the hand information to the occluded region, our HandOccNet reaches the state-of-the-art performance on 3D hand mesh benchmarks that contain challenging hand-object occlusions.
The codes are available in: \href{https://github.com/namepllet/HandOccNet}{https://github.com/namepllet/HandOccNet}.
\end{abstract}

\blfootnote{$^*$ Authors contributed equally.}

\section{Introduction}

\begin{figure}[t]
\begin{center}
\includegraphics[width=0.68\linewidth]{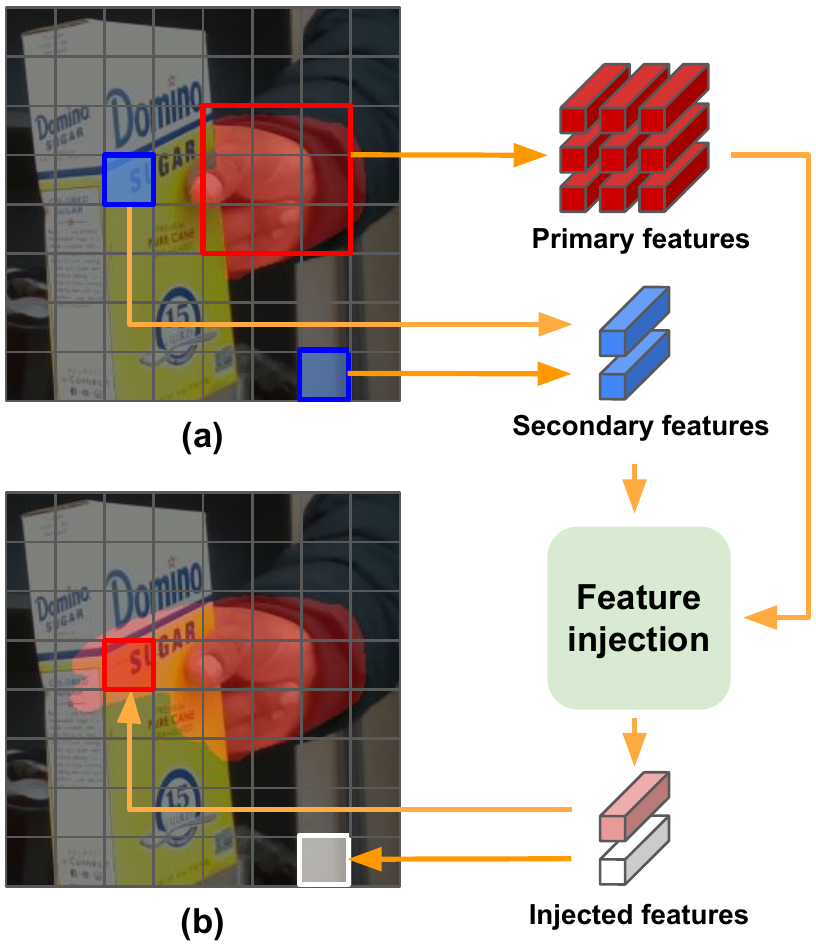}
\end{center}
\vspace*{0mm}
\caption{Example of the operation of the proposed HandOccNet.
(a) The output feature map of spatial attention mechanism for the case of severe occlusion, which consists of sparse primary and secondary features . 
    (b) Our feature injection module finds the primary features related to the secondary features, and then injects the information of primary features into the locations of secondary features.
}
\vspace*{-5mm}
\label{fig:intro_compare}
\end{figure}

Despite promising results of 3D hand mesh estimation from a single RGB image~\cite{ge20193d,choi2020pose2mesh, moon2020i2l,kulon2020weakly,moon2020interhand2,moon2020deephandmesh,moon2020pose2pose,moon2020neuralannot}, making 3D hand mesh estimation method robust to occlusion is still an open challenge. 
One promising approach for the occlusion-robust system is using a spatial attention mechanism.
Although the spatial attention mechanism has not been used for the occlusion-robust 3D hand mesh estimation, several 2D human body pose estimation methods~\cite{chu2017multi,zhu2019robust,zhou2020occlusion} have utilized such attention mechanism for the occlusion-robust results.
They estimate a spatial attention map and multiply it with a feature map to tell the networks where to focus.
The attention map tends to have high scores on human regions and low scores on occluded regions.
Therefore, it attenuates the magnitude of features at occluded regions and makes networks focus on human regions.

Although the spatial attention-based methods have shown noticeable results under the occlusions, there are several limitations.
First, they are mostly for the 2D human body pose estimation, which aims to localize 2D body joint coordinates.
Hence, the validity of their spatial attention mechanism is not proved for the occlusion-robust 3D hand mesh estimation.
In particular, as hands have quite complicated articulations and are often severely occluded by objects, the widely used spatial attention mechanism might fail to produce robust results.
Unlike the methods~\cite{moon2018v2v,moon2017holistic} using a depth map, additional depth ambiguity, which arises from 2D image-to-3D hand estimation, is another bottleneck.
Second, when the occlusions are severe, activations of the spatial attention mechanism become sparse because most of the hand regions are occluded.
The sparse regions contain limited information of hand; hence, relying only on such limited information can lead to erroneous results.

To overcome the above limitations, we propose HandOccNet, a novel framework for occlusion-robust 3D hand mesh estimation.
The main component of the proposed HandOccNet is a feature injection mechanism, shown in Figure~\ref{fig:intro_compare}.
The conventional spatial attention mechanism disregards the information of features at the occluded regions.
On the other hand, our feature injection mechanism utilizes those features as a secondary role to obtain richer representation for occlusion-robust 3D hand mesh estimation.
The ~\textit{primary features} and ~\textit{secondary features} represent features corresponding to high attention scores and low attention scores, respectively.
We leverage information of secondary features to find relevant primary features and inject the information of primary features into the locations of secondary features.
In this process, we use the term ~\textit{inject} to emphasize that the information of secondary features disappears and the information of primary features is injected into empty locations.

To inject not only nearby features but also distant features, we employ Transformer~\cite{vaswani2017attention}, which has an excellent ability to model correlations between features regardless of the distance between features.
Here, the distance between features represents the 2D distance in the pixel space.
We build two Transformer-based modules, feature injecting transformer (FIT) and self-enhancing transformer (SET).
The FIT injects the information of primary features into the regions of the secondary features and outputs a single feature map by utilizing secondary features as queries and primary features as key-value pairs.
The SET utilizes a standard self-attention mechanism to refine the output of the FIT.

Our FIT has two distinctive points compared to the standard Transformer~\cite{vaswani2017attention} for the feature injection. 
First, our FIT computes a correlation map between queries and keys through two types of attention modules, sigmoid-based as well as softmax-based ones, while the standard Transformer uses only softmax-based one.
The softmax-based attention module normalizes the multiplications of each query and all elements of the keys using softmax function.
As softmax considers all elements for the normalization, an undesirable high correlation score can be made when absolute values of all the multiplications are very low but some multiplications are relatively large compared to others.
To prevent such undesirable high correlation scores, we build an additional sigmoid-based attention module.
As the sigmoid activation function does not consider other elements for the normalization, it can avoid the undesired high correlations.
We obtain the final correlation map by multiplying correlation maps from the softmax-based module and sigmoid-based module.
Second, we remove a residual connection between input queries and output of the attention module, while the standard Transformer uses such residual connection.
In other words, the FIT uses queries only when computing correlations between queries and keys and the output feature of FIT does not contain the information of the queries.
This is because we intend secondary features (queries) to be replaced with primary features (values).

We demonstrate the effectiveness of our HandOccNet, through extensive experiments on recently published hand-object interaction datasets, such as HO-3D~\cite{hampali2020honnotate} and FPHA~\cite{garcia2018first}.
These datasets contain various and challenging occlusions in hand regions which reflects realistic occlusions that occur when hands manipulate objects in our daily life.
The experimental results show that our HandOccNet achieves significantly better 3D hand mesh estimation accuracy compared to previous state-of-the-art 3D hand mesh estimation methods.

To summarize, we make the following contributions:
\begin{itemize}
\item We propose a HandOccNet, a novel framework for occlusion-robust 3D hand mesh estimation from a single RGB image.
The proposed HandOccNet utilizes feature injection mechanism that makes feature map robust to occlusion by properly injecting the hand information into the occluded regions.

\item For the feature injection and refinement, we propose two Transformer-based modules, FIT and SET.
The FIT performs the injection mechanism under the guidance of correlations between primary features and secondary features, which represent features of hand regions and occluded regions, respectively.
The SET refines the output feature map of the FIT using a self-attention mechanism.

\item We show our framework significantly outperforms state-of-the-art 3D hand mesh estimation methods on hand-object interaction datasets that contain severe hand occlusions.
\end{itemize}

\section{Related works}

\noindent\textbf{Occlusion-robust human pose estimation.}
There are three main approaches for occlusion-robust human pose estimation.
The first one adopts occlusion-aware data augmentation, the second one leverages temporal information, and the last one utilizes a spatial attention mechanism.

\cite{sarandi2018robust, ke2018multi, cheng20203d} applied occlusion-aware data augmentation in the training time.
Sarandi~\etal~\cite{sarandi2018robust} covered partial region of the image with black solid shapes or object segments from Pascal VOC 2012~\cite{everingham2011pascal} to mimic the occlusions.
Ke~\etal~\cite{ke2018multi} copy background patch of the input image and paste it to human keypoint region.
\cite{cheng20203d,choi20213dcrowdnet} proposed a two stage approach for the 3D pose estimation.
They estimate 2D features for given frames and estimate the 3D pose from the 2D information.
Cheng~\etal~\cite{cheng20203d} utilized sequential 2D features~(2D joint heatmaps) to estimate the consecutive 3D pose.
In training time, \cite{cheng20203d} randomly mask part of estimated 2D joint heatmaps by setting their values to zero in order to simulate occlusions.
The limitation of their augmentations is that the occlusions are synthetic.

\cite{choi2021beyond,cheng2019occlusion} utilized temporal information to compensate for the missing information due to the occlusion.
Choi~\etal~\cite{choi2021beyond} and Cheng~\etal~\cite{cheng2019occlusion} leveraged temporal information for temporally consistent mesh recovery and the occlusion-robust 3D human pose estimation from a video, respectively.
\cite{cheng2019occlusion} first estimated an incomplete 2D pose sequence, which means several joints are labeled as occluded and their coordinates are set to zero, from the input video.
Then they lifted the incomplete 2D pose sequence to complete 3D pose sequence through successive 2D and 3D temporal convolutional networks.

\cite{chu2017multi,zhu2019robust,zhou2020occlusion} utilized spatial attention mechanism for the occlusion-robust system.
Chu~\etal~\cite{chu2017multi} proposed a multi-context, multi-resolution and hierarchical spatial attention scheme for the 2D human pose estimation.
They reweighted the feature map through their spatial attention scheme and boost 2D human pose estimation performance.
Zhu~\etal~\cite{zhu2019robust} first estimated a spatial attention map and multiply it by the feature map to filter out the features of occluded regions.
Then they used inter-feature correlations through a shared structural matrix in order to recover missing features.
Zhou~\etal~\cite{zhou2020occlusion} also estimated the spatial attention map to filter out features of occluded regions. 
Then they recovered features through dilated convolutions.

Ours is related to the spatial attention mechanism; however, there are two main differences compared to the above spatial attention mechanism-based methods.
First, the above methods are mostly designed for 2D human body pose estimation, which is less ambiguous than 3D hand mesh estimation that suffers from depth ambiguity and severe occlusions by objects.
Second, we propose a new feature injection mechanism, which produces highly rich features even when hands are severely occluded.

\noindent\textbf{3D hand mesh estimation under hand-object interaction scenarios.}
After hand object interaction benchmark datasets, such as HO-3D~\cite{hampali2020honnotate} and FPHA~\cite{garcia2018first}, had been released, several studies~\cite{hasson2019learning,hampali2020honnotate,hasson2020leveraging,liu2021semi} have been conducted on these datasets.
Hasson~\etal~\cite{hasson2019learning} proposed novel losses to reflect physical constraints for interacting hand and object. 
Hampali~\etal~\cite{hampali2020honnotate} detected 2D joint locations and fitted a hand model (\textit{i.e.}, MANO~\cite{romero2017embodied}) parameters by minimizing their loss function.
Hasson~\etal~\cite{hasson2020leveraging} leveraged a photometric consistency between neighboring frames.
They estimated mesh for hand and object and rendered it to regress warping flow.
Then they applied a pixel-level loss to enforce photometric consistency between a reference frame and warped frame by the regressed flow.
Liu~\etal~\cite{liu2021semi} proposed a contextual reasoning module that enhances object representations by utilizing interaction between the hand and object.
Most of the above methods focused on modeling interactions between hands and objects.
On the other hand, we firstly introduce a novel feature injection mechanism for the occlusion-robust 3D hand mesh estimation.

\noindent\textbf{Transformers.}
Transformers~\cite{vaswani2017attention} showed superior results on natural language processing (NLP).
Recently, vision researchers have applied Transformers to various applications, such as object detection~\cite{carion2020end}, image classification~\cite{dosovitskiy2020image} and human texture estimation~\cite{xu20213d}.
In the field of 3D human pose and shape estimation,~\cite{huang2020hand,lin2021end,Zheng_2021_ICCV,Yang_2021_ICCV} designed Transformer-based modules.
Huang~\etal~\cite{huang2020hand} proposed Transformer-based networks which estimate 3D hand pose from 3D hand point cloud. 
Lin~\etal~\cite{lin2021end} adopted a Transformer to model global vertex-to-vertex interactions and reconstructed 3D human mesh from a single RGB image.
Zheng~\etal~\cite{Zheng_2021_ICCV} employed spatial and temporal Transformers for 3D human pose estimation in videos.
Yang~\etal~\cite{Yang_2021_ICCV} utilized a Transformer to capture image-specific spatial dependencies between keypoints and estimated 2D human pose. 
Recently, Liu~\etal~\cite{liu2021semi} proposed a Transformer-based contextual reasoning module.
When an object is interacting with a hand in the input image, the contextual reasoning module enhances object regions' features by utilizing hand regions' features.
The enhanced object feature is used only for the 6D object pose estimation, not for the 3D hand mesh estimation.
Liu~\etal~\cite{liu2021semi} is the most relevant work with ours; however, their contextual reasoning module is used only for the 6D object pose estimation.
On the other hand, our injected features are used for the 3D hand mesh estimation.

\section{HandOccNet}

\begin{figure*}[t]
    \centering
    \renewcommand{\wp}{0.99\linewidth}
    \includegraphics[width=\wp]{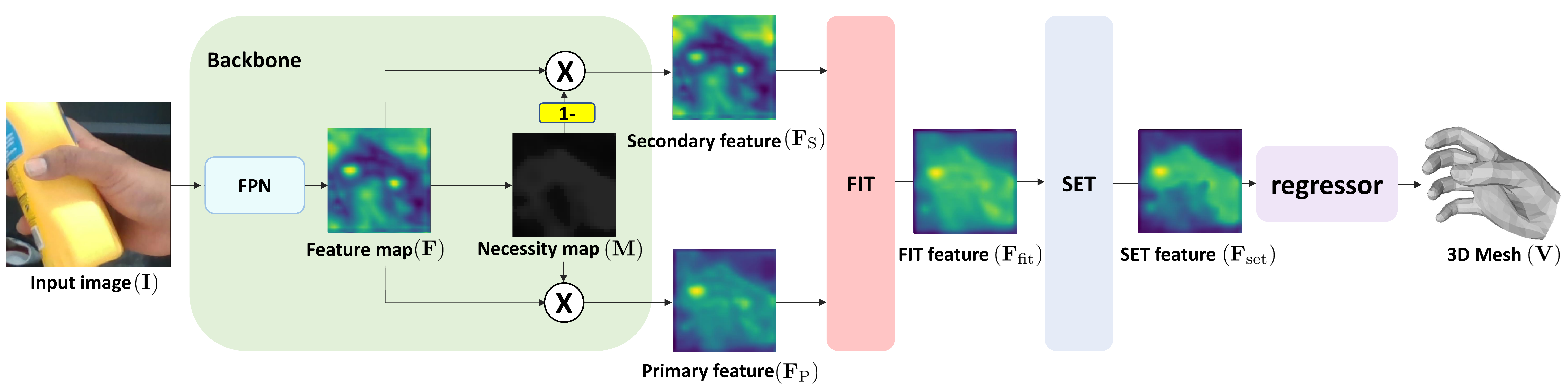}
    \figcspace
    \vspace{-5mm}
    \caption{
    The overall architecure of HandOccNet, which consists of backbone, FIT, SET, and regressor.
    Our HandOccNet extracts primary feature $\mathbf{F}_\text{P}$ and secondary feature $\mathbf{F}_\text{S}$ using a spatial attention mechanism.
    Then, it uses FIT to inject the information of the primary feature $\mathbf{F}_\text{P}$ into the secondary feature $\mathbf{F}_\text{S}$.
    SET refines the output of FIT via self-attention machnism.
    Finally, regressor produces MANO parameters.
    The final 3D hand mesh is obtained by forwarding the MANO parameters to MANO layer.
    The cross mark in a circle represents an element-wise multiplication.
    }
    \label{fig:model}
    \vspace{3mm}
    \figspace
\end{figure*}

\begin{figure}[t]
    \centering
    \renewcommand{\wp}{0.99\linewidth}
    \includegraphics[width=\wp]{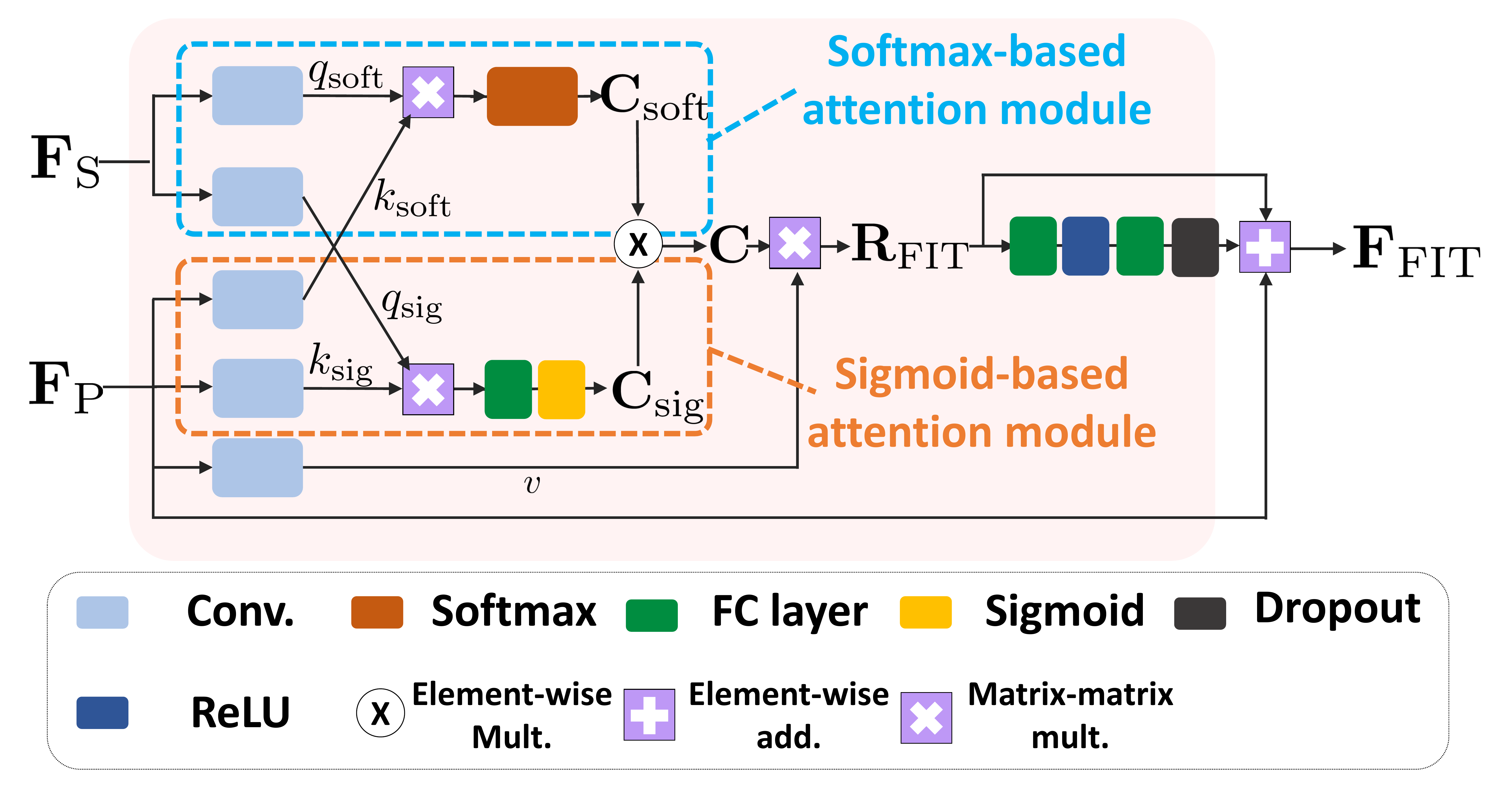}
    \figcspace
    \vspace{-5mm}
    \caption{
    The overall pipeline of FIT.
    FIT injects the primary feature $\mathbf{F}_\text{P}$ into the secondary feature $\mathbf{F}_\text{S}$ using softmax-based attention module and sigmoid-based attention module.
    }
    \label{fig:fit}
    \vspace{3mm}
    \figspace
\end{figure}

In Figure~\ref{fig:model}, we provide an overall pipeline of our HandOccNet for 3D hand mesh estimation.
Our HandOccNet consists of backbone, FIT, SET and regressor.

\subsection{Backbone}
The backbone extracts feature~$\mathbf{F}$ and necessity map$~\mathbf{M}$ from a hand images~$\mathbf{I} \in \mathbb{R}^{512\times 512 \times 3}$.
We first feed the hand image~$\mathbf{I}$ to ResNet50~\cite{he2016deep}-based FPN~\cite{lin2017feature} and resize the output of FPN, which produces a feature map~$\mathbf{F} \in \mathbb{R}^{32 \times 32 \times 256}$.
Then, we obtain a necessity map~$\textbf{M}$ from the feature map~$\mathbf{F}$.
We build three consecutive convolution layers, followed by the sigmoid function to estimate the necessity map~$\textbf{M}$ without supervision so that feature importance could be predicted from learning.
The necessity map~$\mathbf{M}$ represents scores according to spatially varying importance, which is caused by redundant information~(\ie{objects and background}) in feature~$\mathbf{F}$.
Using the necessity map~$\textbf{M}$, we separate the feature map $\mathbf{F}$ into primary feature~$\mathbf{F}_{\text{P}}$ and secondary feature~$\mathbf{F}_{\text{S}}$ with sum-to-one constraints:
\begin{equation*}
    \mathbf{F}_{\text{P}} = \mathbf{F} \otimes \mathbf{M},
    \label{eq:FP}
\end{equation*}
\begin{equation*}
    \mathbf{F}_{\text{S}} = \mathbf{F} \otimes (1-\mathbf{M}).
    \label{eq:FS}
\end{equation*}
$\otimes$ denotes element-wise multiplication.
Note that $\mathbf{F}_{\text{P}}$ contains hand regions' information, which is primarily used for hand mesh estimation and $\mathbf{F}_{\text{S}}$ contains occluded regions' information which is not directly used for hand mesh estimating.
$\mathbf{F}_{\text{P}}$ and $\mathbf{F}_{\text{S}}$ are utilized as query, key, and value for the following FIT.

\subsection{Feature injecting transformer~(FIT)}

The illustration of FIT is shown in Figure~\ref{fig:fit}.
FIT is a Transformer-based module which takes two features, $\mathbf{F}_{\text{P}}$ and $\mathbf{F}_{\text{S}}$, and injects the information of $\mathbf{F}_{\text{P}}$ into $\mathbf{F}_{\text{S}}$ by considering their correlation.
We adopt two sub-modules in the FIT called the softmax-based attention module and sigmoid-based attention module.
The different role of each module is described as follows.

\noindent
\textbf{Softmax-based attention module.}
The softmax-based attention module 
finds the most relevant information of the primary feature $\mathbf{F}_{\text{P}}$ from the secondary feature $\mathbf{F}_{\text{S}}$.
This can be thought as searching for the related hand information in the primary feature $\mathbf{F}_{\text{P}}$ from the occlusion.
Some object information, causing occlusion, can have strong correlation with hand information so that $\mathbf{F}_{\text{S}}$ can tell where to inject the primary feature~$\mathbf{F}_{\text{P}}$.
Therefore, while previous works utilized only $\mathbf{F}_{\text{P}}$ and suppressed $\mathbf{F}_{\text{S}}$ to concentrate on hand information, we use $\mathbf{F}_{\text{S}}$ as a means
of dragging and using $\mathbf{F}_{\text{P}}$.

\begin{figure}[t]
    \captionsetup[subfloat]{font=scriptsize}
    \renewcommand{\wp}{0.22\linewidth}
    \centering
    \subfloat[Input image~($\mathbf{I}$)]{\includegraphics[width=\wp]{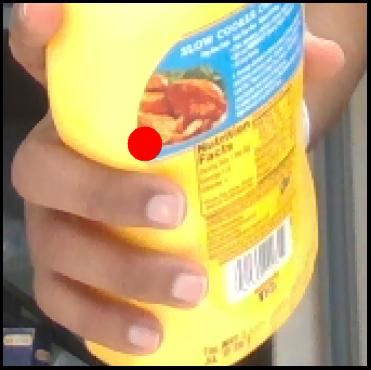}
    \label{fig:img1}
    }
    \subfloat[Multiplication result]{\includegraphics[width=\wp]{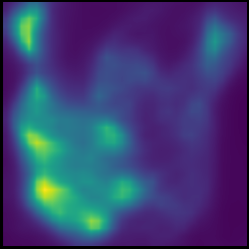}
    \label{fig:dp1}
    }
    \subfloat[$\mathbf{C}_{\text{soft}}$]{\includegraphics[width=\wp]{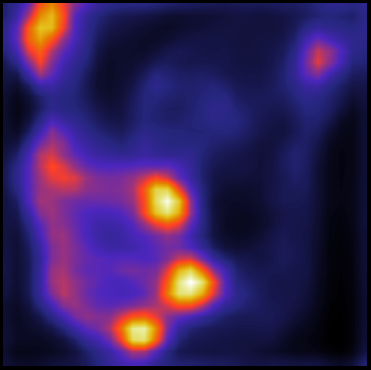}
    \label{fig:soft1}
    }
    \subfloat[$\mathbf{C}$]{\includegraphics[width=\wp]{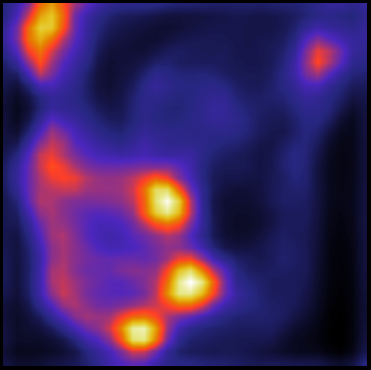}
    \label{fig:softsig1}
    }
    \\
    \subfloat[Input image~($\mathbf{I}$)]{\includegraphics[width=\wp]{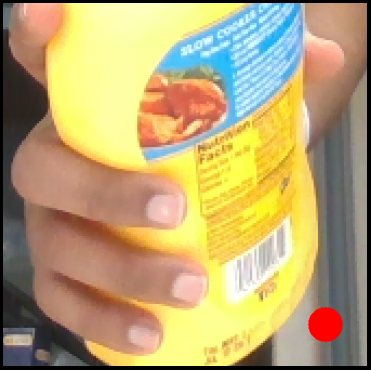}
    \label{fig:img2}
    }
    \subfloat[Multiplication
    result]{\includegraphics[width=\wp]{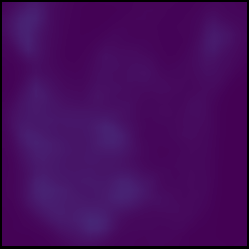}
    \label{fig:dp2}
    }
    \subfloat[$\mathbf{C}_{\text{soft}}$]{\includegraphics[width=\wp]{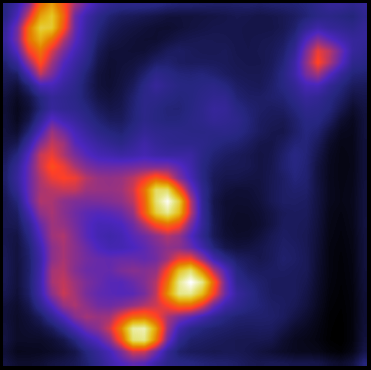}
    \label{fig:soft2}
    }
    \subfloat[$\mathbf{C}$]{\includegraphics[width=\wp]{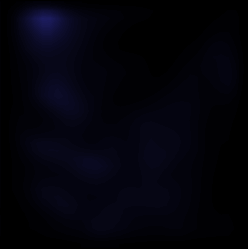}
    \label{fig:softsig2}
    }
    \caption{
    (a) and (e): red points represent example locations of query $q_\text{soft}$ overlayed on the input image. 
    (b) and (f): multiplication between the red points of query $q_\text{soft}$ (shown in (a) and (e), respectively) and all elements of key $k_\text{soft}$.
    (c) and (g): $\mathbf{C}_{\text{soft}}$ calculated from (b) and (f), respectively, by applying a softmax function.
    (d) and (h) : $\mathbf{C}$ calculated from element-wise multiplication of sigmoid-based correlation map~$\mathbf{\text{sig}}$ and $\mathbf{C}_{\text{soft}}$.
    }
    \label{fig:softmaxvisual}
    \vspace{3mm}
    \figspace
\end{figure}

We extract query~$q_{\text{soft}}$ from $\mathbf{F}_{\text{S}}$ and key~$k_\text{soft}$ from $\mathbf{F}_{\text{P}}$ by two $1 \times 1$ convolution layer.
Then we reshape the query and key to dimension~$\mathbb{R}^{1024 \times 256}$, where 1024 represents the multiplication of width and height of $\mathbf{F}_{\text{P}}$ and $\mathbf{F}_{\text{S}}$.
By recalling the attention mechanism of the previous Transformers~\cite{liu2021semi, vaswani2017attention, dosovitskiy2020image}, 
the softmax-based attention module generates the correlation map $\mathbf{C}_{\text{soft}} \in \mathbb{R}^{1024 \times 1024}$ from the softmax function after the matrix multiplication of query~$q_\text{soft}$ and key~$k_\text{soft}$:
\begin{equation}
    \textbf{C}_{\text{soft}} = \text{softmax}(\frac{{q_{\text{soft}}}{k_{\text{soft}}}^{T}}{\sqrt{d_{k_{\text{soft}}}}}),
    \label{eq:soft}
\end{equation}
where $d_{k_{\text{soft}}}=256$ denotes the feature dimension of the key~$k_{\text{soft}}$.
The correlation map~$\textbf{C}_{\text{soft}}$ indicates how much information is related between each pixel of query~$q_{\text{soft}}$ and key~$k_{\text{soft}}$.
In other words, $\textbf{C}_{\text{soft}}$ can be utilized to find which information of $\mathbf{F}_{\text{P}}$
to use to fill the information of $\mathbf{F}_{\text{S}}$.
However, using only softmax for the activation is limited in handling correlation when the overall key information is not related to the specific query pixel.
For example, some information~(\ie{background}) in secondary feature~
$\mathbf{F}_{\text{S}}$ can be not related to the overall $\mathbf{F}_{\text{P}}$ as in Figure~\ref{fig:img2} so that the multiplication result before the softmax function might show low values for all elements of key~$k_{\text{soft}}$ as shown in Figure~\ref{fig:dp2}.
Nevertheless, the softmax function approximates an absolutely small number, which is relatively larger than others, to a high score.
Therefore, as shown in Figure~\ref{fig:soft2}, undesired high correlation can occur from some relatively high elements, which are absolutely low.
To use only the advantages seen in Figure~\ref{fig:soft1}, which properly displays the correlation based on high multiplication result~\ref{fig:dp1}, and handle the problems shown in Figure~\ref{fig:soft2}, we build an additional sigmoid-based attention module to filter the undesired high correlation score.

\noindent
\textbf{Sigmoid-based attention module.}
The sigmoid-based attention module filters the undesired high correlation by generating a correlation map between each query pixel and the global key information.
We extract additional key-query pair, $k_{\text{sig}}$ and $q_{\text{sig}}$, with same process of extracting $k_{\text{soft}}$ and $k_{\text{soft}}$.
Then, the module generates the correlation map~$\mathbf{C}_{\text{sig}} \in \mathbb{R}^{1024 \times 1}$ as follows:
\begin{equation}
    \mathbf{C}_{\text{sig}} = \text{sigmoid}(\text{pool}(\frac{{q_{\text{sig}}}{k_{\text{sig}}}^{T}}{\sqrt{{d_{k_\text{sig}}}}})),
    \label{eq:sig}
\end{equation}
where pool denotes average pooling to aggregate correlation between each query~$q_{\text{sig}}$ and all elements of key~$k_{\text{sig}}$.
$d_{k_{\text{sig}}}=256$ denotes the feature dimension of the key~$k_{\text{sig}}$.
The average pooling along the key dimension can make the correlation map $\mathbf{C}_\text{sig}$ robust to noisy correlations.
We observed that removing the pooling in our sigmoid-based attention module makes our HandOccNet diverge during the training.

Unlike softmax function, which normalizes input element to a probability distribution considering the other elements of input, sigmoid only concentrates on normalizing a single element to a probability.
Therefore, the sigmoid function does not suffer from the undesired high correlation problem of the softmax function by producing small attention scores from the small numbers of the multiplication result.
We obtain our final correlation map~$\textbf{C} \in \mathbb{R}^{1024 \times 1024}$ by using both correlation map from sigmoid and softmax based module, $\mathbf{C}_{\text{soft}}$ and $\mathbf{C}_{\text{sig}}$ like below:
\begin{equation*}
    \textbf{C} = \textbf{C}_{\text{soft}} \otimes \textbf{C}_{\text{sig}}.
    \label{eq:correlation}
\end{equation*}
Figure~\ref{fig:dp2}, ~\ref{fig:soft2}, and ~\ref{fig:softsig2} show the effectiveness of the correlation map from the sigmoid-based attention module.
Figure~\ref{fig:soft2} shows high correlations although Figure~\ref{fig:dp2} has small multiplication results, which represents that Figure~\ref{fig:soft2} suffers from the undesired high correlations.
By multiplying $\textbf{C}_{\text{sig}}$ to Figure~\ref{fig:soft2}, we fix the undesired high correlations, as shown in Figure~\ref{fig:softsig2}.

\noindent
\textbf{Feature injection.}
Using the correlation map $\mathbf{C}$, we inject the hand information to the proper occluded region.
Please note that we use the word ``injection" because, unlike typical Transformers~\cite{vaswani2017attention} that use query information in output with residual connection, the query information disappears and the information of value is injected into the empty locations.
We get value $v \in \mathbb{R}^{1024 \times 256}$, which represents the source information indexed by the keys in Transformer, from $\mathbf{F}_{\text{P}}$ with a 1x1 convolution and flattening its spatial dimension.
Then, we inject the value into the low importance region to obtain a residual feature $\mathbf{R}_\text{FIT} \in \mathbb{R}^{1024 \times 256}$ like below:
\begin{equation}
    \mathbf{R}_{\text{FIT}} = \mathbf{C}v.
    \label{eq:finres}
\end{equation}
Afterward, we feed $\mathbf{R}_\text{FIT}$ into a feed-forward module.
The feed-forward module consists of a two-layer MLP and layer normalization with a residual connection between its input and output.
We further add a residual connection between its output and the primary feature~$\mathbf{F}_\text{P}$, which already contains essential information for hand mesh estimation.
FIT's output feature $\mathbf{F}_\text{FIT} \in \mathbb{R}^{32 \times 32 \times 256}$ is obtained like below:
\begin{equation*}
    \mathbf{F}_{\text{FIT}} = \mathbf{F}_{\text{P}} + \psi(\mathbf{R}_{\text{FIT}}) + \psi(\text{MLP}(\text{LN}(\mathbf{R}_{\text{FIT}}))),
    \label{eq:finfinal}
\end{equation*}
where $\psi$ denotes a reshaping function that reshapes the input feature to $\mathbb{R}^{32 \times 32 \times 256}$. 
\text{MLP} and \text{LN} denote the MLP module and layer normalization layer, respectively.

\begin{figure}[t]
    \centering
    \renewcommand{\wp}{0.99\linewidth}
    \includegraphics[width=\wp]{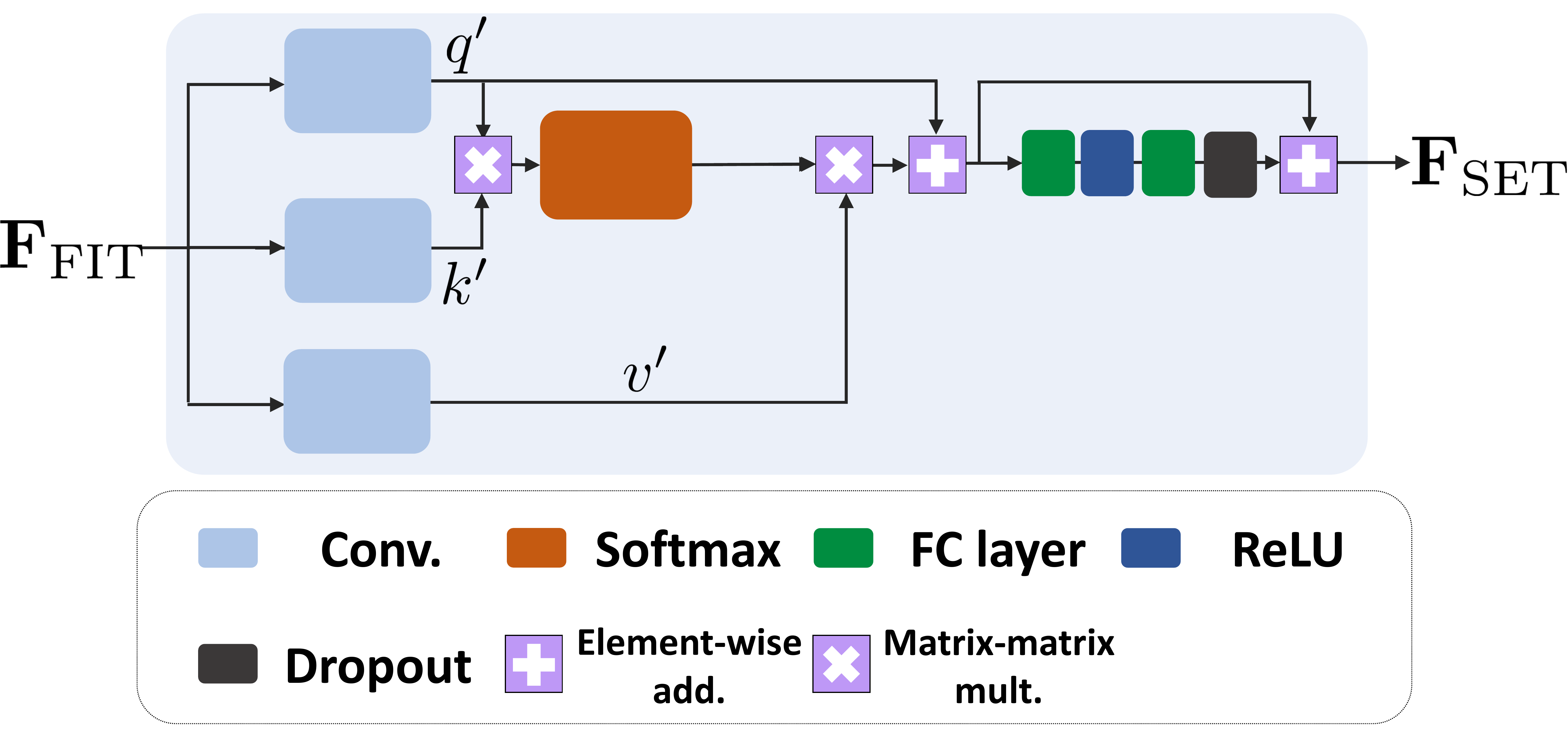}
    \figcspace
    \vspace{-5mm}
    \caption{
    The overall pipeline of SET.
    SET refines the feature $\mathbf{F}_\text{FIT}$ with self-attention mechanism.
    }
    \label{fig:set}
    \vspace{3mm}
    \figspace
\end{figure}

\subsection{Self-Enhancing transformer~(SET)}
The illustration of SET is shown in Figure~\ref{fig:set}.
SET refines the feature~$\mathbf{F}_{\text{FIT}}$ by referencing the distant information from feature~$\mathbf{F}_{\text{FIT}}$ with self-attention.
Different from the FIT which concentrates on injecting primary feature~$\mathbf{F}_{\text{P}}$ into secondary feature~$\mathbf{F}_{\text{S}}$, 
SET utilizes self-attention of $\mathbf{F}_{\text{FIT}}$ by extracting the query~$q'$, key~$k'$, and value~$v'$ from the same feature~$\mathbf{F}_{\text{FIT}}$ with three 1x1 convolution layers.
As SET performs self-attention, there is no existence of case that overall key information is not related to the query pixel because each query pixel is at least correlated to itself.
Therefore, instead of using the sigmoid-based attention module which is used to filter the undesired high correlation, we only adopt the softmax-based attention module to obtain a correlation map in SET.
SET follows the same pipeline of the softmax-based attention module in FIT except a residual connection between query $q'$ and the multiplication of correlation map and value $v'$.
The module in FIT does not have the residual connection as its goal is to ``replace" the query with the value for the feature injection.
On the other hand, as the goal of SET is enhancing the input feature, not the injection, we add the residual connection, following previous Transformers~\cite{vaswani2017attention}.
The output of SET is denoted by $\mathbf{F}_\text{SET}$.
Two or more SET do not have much effect in our experiment because sufficient enhancement is already occurred in the first SET; therefore, we use one SET after the FIT.

\subsection{Regressor}
The regressor produces MANO pose and shape parameters, and the final 3D hand mesh is obtained by forwarding the MANO parameters to MANO layer.
First, a single block of hourglass network~\cite{newell2016stacked} takes enhanced feature $\mathbf{F}_{\text{SET}}$ as input and outputs 2D heatmaps for each joint $\mathbf{H}$.
Then, four residual blocks~\cite{he2016deep} takes a concatenation of the enhanced hand feature~$\mathbf{F}_{\text{SET}}$ and the 2D heatmap $\mathbf{H}$.
Finally, the output of the residual blocks are vectorized into a 2048 dimensional vector and passed to fully-connected layers, which predict MANO pose parameters~$\theta \in \mathbb{R}^{48}$ and shape parameters~$\beta \in \mathbb{R}^{10}$.
\fix{We multiplied the joint regression matrix to a 3D mesh in rest pose and applied the forward kinematics to get the final 3D hand joints coordinates and obtained the final 3D hand mesh~$\mathbf{V} \in \mathbb{R}^{778 \times 3}$.}

To train our HandOccNet, we minimize a loss function, defined as a combination of L2 distances between the predicted and ground truths $\mathbf{H}$, $\theta$, $\beta$, $\mathbf{V}$, and $J^\text{3D}$.
$J^\text{3D}$ denotes a 3D hand joint coordinates, obtained by multiplying a joint regression matrix to 3D hand mesh $\mathbf{V}$, where the matrix is defined in MANO.

\section{Experiments}

\subsection{Implementation details}
All implementations were done with PyTorch~\cite{paszke2019pytorch}.
We use Adam optimizer~\cite{kingma2014adam} with batch size 24 for our training.
On HO-3D and FPHA, each model was trained with annealing the learning rate at every 10th from the initial learning rate $10^{-4}$.
All other details will be available in our codes.

\subsection{Datasets and evaluation metrics}
\noindent
\textbf{HO-3D.}
The HO-3D dataset~\cite{hampali2020honnotate} is a hand-object interaction dataset which contains challenging occlusions.
This dataset provides RGB images with MANO-based hand joints and meshes, and camera parameters.
The results on the test set can be evaluated via an online submission system.


\begin{table}[t]
    \newcommand{\cmark}{\ding{51}}%
    \newcommand{\xmark}{\ding{55}}%
    \tabcspace
    \centering
    \resizebox{0.8\linewidth}{!}{
    \begin{tabular}{c|cccc}
        \toprule
        Architectures & Joint & Mesh & F@5 & F@15 \\
        \midrule
        Identity & 10.6 & 10.0 & 52.5 & 94.9 \\
        Residual blocks & 10.2 & 9.8 & 51.0 & 95.3 \\
        FIT & 9.4 & 9.2 & 54.3 & 96.0 \\
        SET & 9.8 & 9.6 & 52.6 & 95.3 \\
        \textbf{FIT + SET (Ours)} & \textbf{9.1} & \textbf{8.8} & \textbf{56.4} & \textbf{96.3} \\
        \bottomrule
    \end{tabular}}
    \caption{Comparison of models with various architectures on HO-3D.}
    \label{tab:modelablation}
    \tabspace
    \vspace{3mm}
\end{table}
\begin{figure}[t]
    \captionsetup[subfloat]{font=scriptsize}
    \renewcommand{\wp}{0.27\linewidth}
    \centering
    \includegraphics[width=\wp]{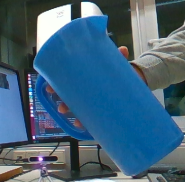}
    \hfill
    \includegraphics[width=\wp]{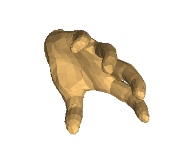}
    \hfill
    \includegraphics[width=\wp]{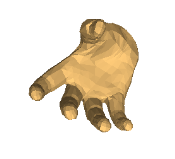}\\
	\vspace{-3mm}
    \subfloat[Input image~($\mathbf{I}$)]{\includegraphics[width=\wp]{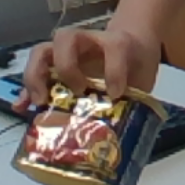}}
    \hfill
    \subfloat[Wo. FIT and SET]{\includegraphics[width=\wp]{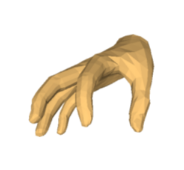}}
    \hfill
    \subfloat[W. FIT and SET~(Ours)]{\includegraphics[width=\wp]{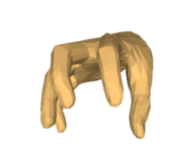}}
    \caption{Comparisons between models without and with FIT and SET on HO-3D.
    }
    \label{fig:baselinehand}
    \vspace{3mm}
    \figspace
\end{figure}

\noindent
\textbf{First-Person Hand Action (FPHA).}
The FPHA dataset~\cite{garcia2018first} contains egocentric RGB-D videos capturing a wide range of hand-object interactions.
While 3D hand pose annotations are available in all frames, 6D object pose annotations are available in a small subset of the entire dataset.
For the fair comparison, we follow the same train and test set split as previous works~\cite{hasson2020leveraging,liu2021semi}.


\noindent
\textbf{Evaluation metrics.}
For HO3D, we report the standard metrics, such as mean joint error and mesh error in mm and F-scores, returned from the official evaluation server.
For FPHA, we report the mean joint error in mm.
All metrics are obtained after the procrustes alignment.
\fix{Furthermore, as results before procrustes alignment are also important, we also show joint error before procrsutes alignment on the HO3D dataset in the supplementary material.}
\begin{figure}[t]
    \captionsetup[subfloat]{font=footnotesize}
    \renewcommand{\wp}{0.34\linewidth}
    \centering
    \subfloat[Input image~($\mathbf{I}$)]{\includegraphics[width=\wp]{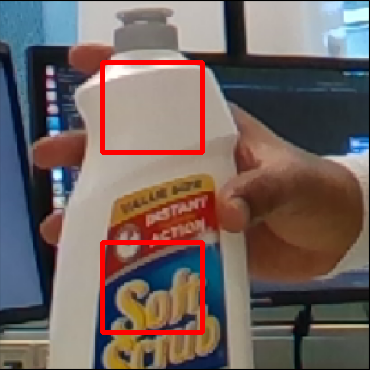}
    \label{fig:featureflow1}
    }
    \subfloat[Primary feature~($\mathbf{F}_{\text{P}}$)]{\includegraphics[width=\wp]{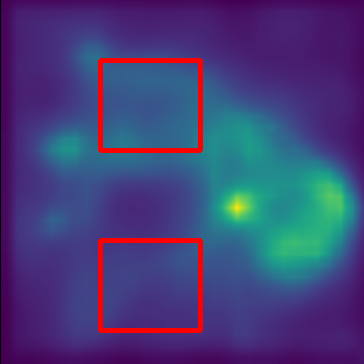}
    \label{fig:featureflow2}
    }\\
    \subfloat[Output of FIT~($\mathbf{F}_{\text{FIT}}$)]{\includegraphics[width=\wp]{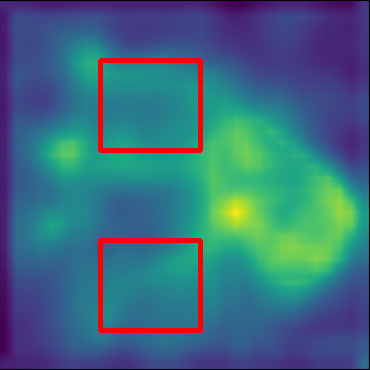}
    \label{fig:featureflow3}
    }
    \subfloat[Output of SET~($\mathbf{F}_{\text{SET}}$)]{\includegraphics[width=\wp]{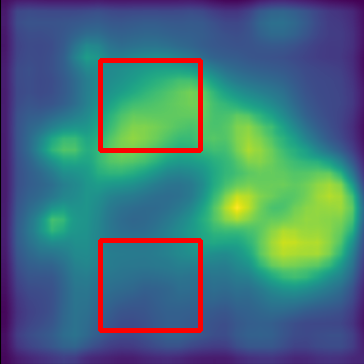}
    \label{fig:featureflow4}
    }
    \caption{Visualization of the feature map. Our FIT successfully injects information into the occluded region and SET makes richer information in the occluded region by self-enhancing.
    }
    \label{fig:featureflow}
    \vspace{2mm}
    \figspace
\end{figure}

\subsection{Ablation study}

\noindent\textbf{FIT and SET.}
Table~\ref{tab:modelablation} shows that using our FIT and SET consistently improves all metrics, which demonstrates their benefits.
Figure~\ref{fig:baselinehand} further shows that FIT and SET improve the accuracy of 3D hand mesh when severe occlusions are included in the input image.
For the comparison, we design four variants.
All the variants have the same backbone and regressor, shown in Figure~\ref{fig:model}, and different components between the backbone and regressor.
The first and second ones have a similar pipeline with that of conventional spatial attention mechanism.
The first one passes the primary feature directly to the regressor, and the second one passes the primary feature to six residual blocks~\cite{he2016deep} without introducing any Transformer-based modules.
They produce worse results than ours, which indicates that our newly introduced feature injection mechanism using two Transformers is highly beneficial.
The third and fourth variants solely use one of FIT and SET, which produce worse results than ours.
This demonstrates the efficacy of architecture of our HandOccNet using a combination of both FIT and SET.

Figure~\ref{fig:featureflow} shows how our FIT enhances the feature of occluded regions.
Initially, red boxes in Figure~\ref{fig:featureflow2} lack hand information due to the occlusion.
Then, FIT injects hand information into the occluded region, which results in solid activation at the occluded region (red boxes), as shown in Figure~\ref{fig:featureflow3}.
Furthermore, SET enhances the information to obtain richer representation for occlusion-robust 3D hand mesh estimation, as shown in Figure~\ref{fig:featureflow4}.

\noindent
\textbf{Architecture of FIT.}
Table~\ref{tab:sigmoideablation} shows that our combination of softmax-based and sigmoid-based attention modules in FIT achieves the best results in all metrics.
The sigmoid-based one filters the undesired high correlation, as shown in Figure~\ref{fig:softmaxvisual}.
Compared to ours, using only softmax-based one like standard Transformer suffers from the undesired high correlation, which results in worse results.
We also report the results of a combination of two softmax-based ones.
This combination produces worse results than using a single softmax-based one, which indicates simply stacking the softmax-based ones cannot fix the undesired high correlations.

\begin{table}[t]
    \newcommand{\cmark}{\ding{51}}%
    \newcommand{\xmark}{\ding{55}}%
    \tabcspace
    \centering
    \resizebox{1.0\linewidth}{!}{
    \begin{tabular}{c|cccc}
        \toprule
        FIT architectures & Joint & Mesh & F@5 & F@15 \\
        \midrule
        Softmax attn. & 9.5 & 9.1 & 54.5 & 95.9\\
        Softmax attn. + Softmax attn. & 9.6 & 9.2 & 53.6 & 95.9\\
        \textbf{Softmax attn. + Sigmoid attn. (Ours)} & \textbf{9.1} & \textbf{8.8} & \textbf{56.4} & \textbf{96.3}\\
        \bottomrule
    \end{tabular}}
    \caption{Comparison of models with various FIT architectures on HO-3D.}
    \label{tab:sigmoideablation}
    \tabspace
\end{table}
\begin{table}[t]
    \newcommand{\cmark}{\ding{51}}%
    \newcommand{\xmark}{\ding{55}}%
    \tabcspace
    \centering
    \resizebox{1.0\linewidth}{!}{
    \begin{tabular}{c|cccc}
        \toprule
        Settings & Joint & Mesh & F@5 & F@15 \\
        \midrule
        Residual connection with $q_{\text{soft}}$ & 9.5 & 9.1 & 55.0 & 96.0 \\
        Residual connection with $q_{\text{sig}}$ & 9.7 & 9.3 & 53.3 & 95.7 \\
        \textbf{Without residual connections (Ours)} & \textbf{9.1} & \textbf{8.8} & \textbf{56.4} & \textbf{96.3} \\
        \bottomrule
    \end{tabular}}
    \caption{Comparison between models that have and do not have residual connections with query in FIT on HO-3D.}
    \label{tab:residualconnection}
    \tabspace
\end{table}
\begin{table}[t]
    \newcommand{\cmark}{\ding{51}}%
    \newcommand{\xmark}{\ding{55}}%
    \tabcspace
    \centering
    \resizebox{1.0\linewidth}{!}{
    \begin{tabular}{c|cccc}
        \toprule
        SET architectures & Joint & Mesh & F@5 & F@15 \\
        \midrule
        Identity & 9.4 & 9.2 & 54.3 & 96.0 \\
        Residual blocks & 9.6 & 9.2 & 54.4 & 95.9\\
        \textbf{Single Transformer (Ours)} & 
        \textbf{9.1} & \textbf{8.8} & \textbf{56.4} & \textbf{96.3}\\
        Two Transformers & 9.2 & 8.9 & 56.2 & \textbf{96.3} \\
        \bottomrule
    \end{tabular}}
    \caption{Comparison of models with various SET architecture on HO-3D.}
    \label{tab:setablation}
    \tabspace
\end{table}


\begin{figure*}[t]
\begin{center}
\includegraphics[width=0.72\linewidth]{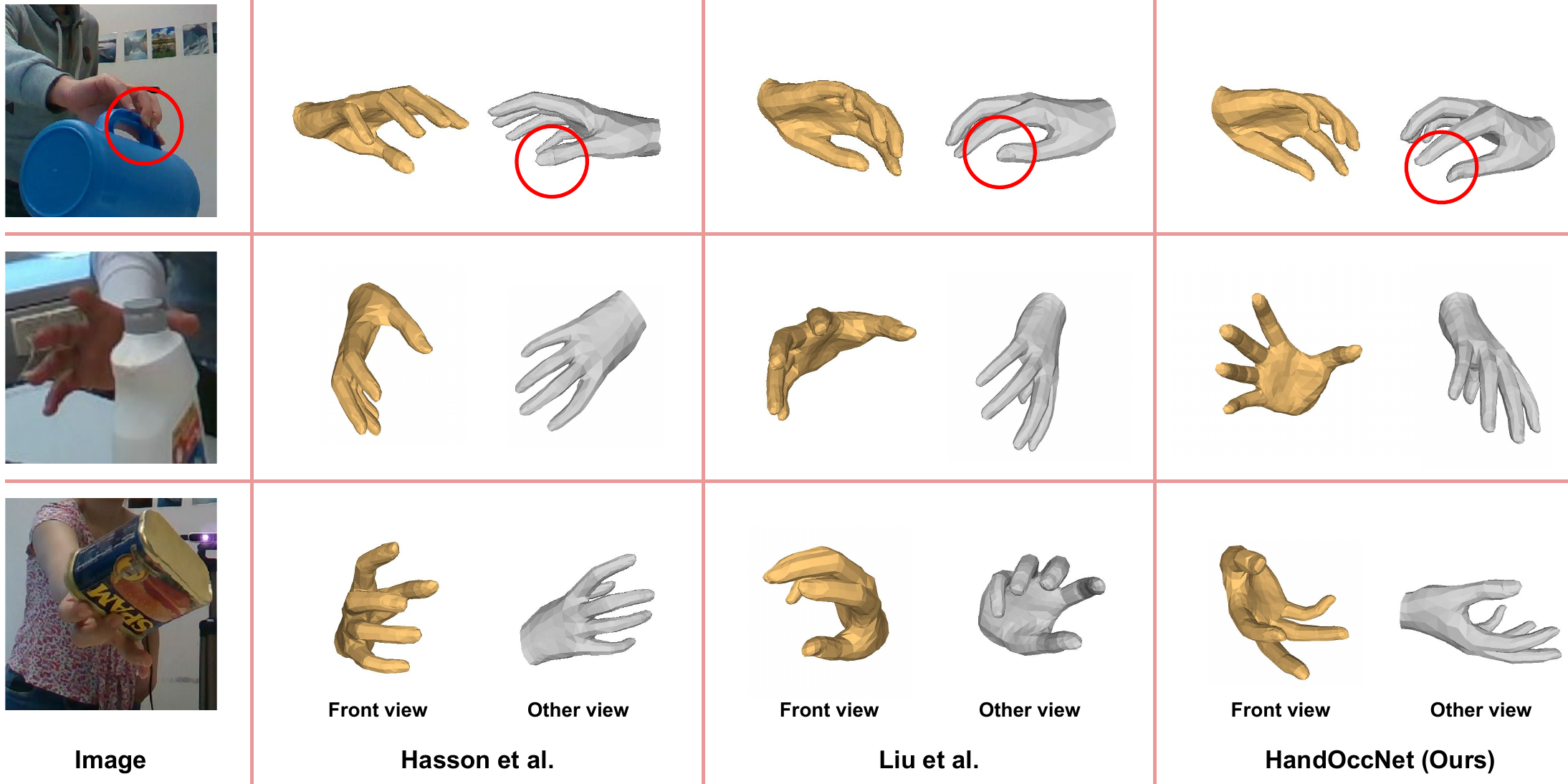}
\end{center}
\vspace*{-7mm}
\caption{
Qualitative comparison of the proposed HandOccNet and state-of-the-art 3D hand mesh estimation methods~\cite{hasson2020leveraging,liu2021semi} on HO-3D.}
\vspace*{-7mm}
\label{fig:results}
\end{figure*}

\noindent
\textbf{Feature injection in FIT.}
Table~\ref{tab:residualconnection} shows that removing the two residual connections achieves the best results.
The first residual connection is a connection between the query of the softmax-based attention module $q_{\text{soft}}$ and the residual feature ${\mathbf{R}}_{\text{FIT}}$.
The second one is a connection between the query of the sigmoid-based attention module $q_{\text{sig}}$ and the residual feature ${\mathbf{R}}_{\text{FIT}}$.
Unlike standard Transformers, our FIT does not have the residual connection between a query and residual feature, which is a multiplication of the correlation map and value (see Eq.~\ref{eq:finres}).
This is because our FIT is designed to ``inject" the information of value into the location of query; therefore, query is used only for the correlation map computation (see Eq.~\ref{eq:soft} and ~\ref{eq:sig}).
The comparisons show that the residual connections are harmful for the feature injection, which results in worse performance.

\begin{table}[t]
    \newcommand{\cmark}{\ding{51}}%
    \newcommand{\xmark}{\ding{55}}%
    \tabcspace
    \centering
    \resizebox{0.7\linewidth}{!}{
    \begin{tabular}{c|cccc}
        \toprule
        Methods & Joint & Mesh & F@5 & F@15 \\
        \midrule
        Pose2Mesh~\cite{choi2020pose2mesh} & 12.5  & 12.7 & 44.1 & 90.9 \\
        Hasson~\etal~\cite{hasson2020leveraging} & 11.4  & 11.4 & 42.8 & 93.2\\
        I2L-MeshNet~\cite{moon2020i2l} & 11.2  & 13.9 & 40.9 & 93.2 \\
        Hasson~\etal~\cite{hasson2019learning} & 11.1 & 11.0 & 46.0 & 93.0\\
        Hampali~\etal~\cite{hampali2020honnotate} & 10.7 &  10.6 & 50.6 & 94.2\\
        METRO~\cite{lin2021end} & 10.4 &  11.1 & 48.4 & 94.6 \\
        Liu~\etal~\cite{liu2021semi} & 10.2 &  9.8 & 52.9 & 95.0 \\
        \textbf{HandOccNet (Ours)} & \textbf{9.1} &  \textbf{8.8} & \textbf{56.4} & \textbf{96.3}\\
        \bottomrule
    \end{tabular}}
        \vspace{-3mm}
    \caption{Comparison with state-of-the-art methods on HO-3D. PA denotes Procrustes Alignment.}
    \label{tab:ho3d}
    \tabspace
\end{table}

\noindent
\textbf{Architecture of SET.}
Table~\ref{tab:setablation} shows that designing SET as a single Transformer achieves the best results, which validates our design choice of SET.
For the demonstration, we design three variants that have different SET architectures.
The first one does not introduce any learnable modules in SET and just set its input feature $\mathbf{F}_\text{FIT}$ to the output feature $\mathbf{F}_\text{SET}$.
The comparison with ours shows that the absence of learnable modules in SET produce worse results than ours, which indicates that additional feature processing is necessary.
The second one uses a series of local feature extractor, which consists of three residual blocks~\cite{he2016deep}.
The comparison shows that adding such local feature extractors produces worse results than ours and even worse than the first variant that does not introduce any learnable modules.
This is because the newly injected features in the input feature $\mathbf{F}_\text{FIT}$ are not locally associated.
As the feature injection is performed by Transformers in FIT, distant features can be injected.
Therefore, the injected features can have very different information from features of nearby pixels.
Due to such locally non-associated features, the local feature extractors can have difficulty in learning local patterns, which results in worse performance.
The third one uses two Transformers, which achieves slightly worse results than our single Transformer-based module.
This is because a single Transformer already enhances the feature sufficiently so that the additional Transformer has a marginal effect on enhancing the input feature.

\subsection{Comparisons with the state-of-the-art methods}

\begin{table}[t]
    \newcommand{\cmark}{\ding{51}}%
    \newcommand{\xmark}{\ding{55}}%
    \tabcspace
    \centering
    \resizebox{0.7\linewidth}{!}{
    \begin{tabular}{c|c}
        \toprule
        Methods & 3D joint error \\
        \midrule
        I2L-MeshNet~\cite{moon2020i2l} & 21.2 \\
        Hasson~\etal~\cite{hasson2020leveraging} & 18.0\\
        Liu~\etal~\cite{liu2021semi} & 16.0\\
        Hasson~\etal~\cite{hasson2019learning} & 14.9\\
        \textbf{HandOccNet (Ours)} & \textbf{10.8} \\
        \bottomrule
    \end{tabular}}
        \vspace{-3mm}
    \caption{Comparison with state-of-the-art methods on FPHA.}
    \label{tab:fpha}
    \tabspace
\end{table}

Table~\ref{tab:ho3d} and ~\ref{tab:fpha} show that our HandOccNet achieves the best results on HO-3D and FPHA, respectively.
Figure~\ref{fig:results} shows that our HandOccNet produces much better results than state-of-the-art methods on HO-3D.
As shown in the figure, our HandOccNet estimates global rotation of the hand accurately, even under the severe occlusion.
Overall, our HandOccNet outperforms state-of-the-art methods on HO-3D and FPHA, which contain diverse hand-object occlusions.
The results are consistent with the ablation study, which shows the proposed feature injection mechanism.
\fix{Moreover, we show comparisons on larger dataset, Dex-YCB~\cite{chao2021dexycb}, to justify the efficacy of our HandOccNet in supplementary material.}

\section{Conclusion}
We present HandOccNet, a novel 3D hand mesh estimation framework that is robust to occlusions.
Our HandOccNet utilizes a feature injection mechanism that makes feature map robust to occlusion by properly injecting the information of primary features into the location of secondary features.
To this end, we design two successive Transformers: FIT and SET.
Our experimental results show that our method achieves the state-of-the-art performance on 3D hand mesh benchmarks that contain severe occlusions.
\section{Acknowledgment}
This work was supported in part by IITP grant funded by the Korea government (MSIT) [No. 2021-0-01343, Artificial Intelligence Graduate School Program (Seoul National University)].
\clearpage


\bibliography{main}

\begin{thebibliography}{10}\itemsep=-1pt

\bibitem{carion2020end}
Nicolas Carion, Francisco Massa, Gabriel Synnaeve, Nicolas Usunier, Alexander
  Kirillov, and Sergey Zagoruyko.
\newblock End-to-end object detection with transformers.
\newblock In {\em ECCV}, 2020.

\bibitem{chao2021dexycb}
Yu-Wei Chao, Wei Yang, Yu Xiang, Pavlo Molchanov, Ankur Handa, Jonathan
  Tremblay, Yashraj~S Narang, Karl Van~Wyk, Umar Iqbal, Stan Birchfield, et~al.
\newblock {DexYCB}: A benchmark for capturing hand grasping of objects.
\newblock In {\em CVPR}, 2021.

\bibitem{cheng20203d}
Yu Cheng, Bo Yang, Bo Wang, and Robby~T Tan.
\newblock {3D} human pose estimation using spatio-temporal networks with
  explicit occlusion training.
\newblock In {\em AAAI}, 2020.

\bibitem{cheng2019occlusion}
Yu Cheng, Bo Yang, Bo Wang, Wending Yan, and Robby~T Tan.
\newblock Occlusion-aware networks for {3D} human pose estimation in video.
\newblock In {\em ICCV}, 2019.

\bibitem{choi2021beyond}
Hongsuk Choi, Gyeongsik Moon, Ju~Yong Chang, and Kyoung~Mu Lee.
\newblock Beyond static features for temporally consistent {3D} human pose and
  shape from a video.
\newblock In {\em CVPR}, 2021.

\bibitem{choi2020pose2mesh}
Hongsuk Choi, Gyeongsik Moon, and Kyoung~Mu Lee.
\newblock {Pose2Mesh}: Graph convolutional network for {3D} human pose and mesh
  recovery from a {2D} human pose.
\newblock In {\em ECCV}, 2020.

\bibitem{choi20213dcrowdnet}
Hongsuk Choi, Gyeongsik Moon, JoonKyu Park, and Kyoung~Mu Lee.
\newblock Learning to estimate robust {3D} human mesh from in-the-wild crowded
  scenes.
\newblock In {\em CVPR}, 2022.

\bibitem{chu2017multi}
Xiao Chu, Wei Yang, Wanli Ouyang, Cheng Ma, Alan~L Yuille, and Xiaogang Wang.
\newblock Multi-context attention for human pose estimation.
\newblock In {\em CVPR}, 2017.

\bibitem{dosovitskiy2020image}
Alexey Dosovitskiy, Lucas Beyer, Alexander Kolesnikov, Dirk Weissenborn,
  Xiaohua Zhai, Thomas Unterthiner, Mostafa Dehghani, Matthias Minderer, Georg
  Heigold, Sylvain Gelly, Jakob Uszkoreit, and Neil Houlsby.
\newblock An image is worth 16x16 words: Transformers for image recognition at
  scale.
\newblock In {\em ICLR}, 2021.

\bibitem{everingham2011pascal}
Mark Everingham and John Winn.
\newblock The pascal visual object classes challenge 2012 ({VOC}2012)
  development kit.
\newblock In {\em PASCAL}, 2011.

\bibitem{garcia2018first}
Guillermo Garcia-Hernando, Shanxin Yuan, Seungryul Baek, and Tae-Kyun Kim.
\newblock First-person hand action benchmark with {RGB-D} videos and {3D} hand
  pose annotations.
\newblock In {\em CVPR}, 2018.

\bibitem{ge20193d}
Liuhao Ge, Zhou Ren, Yuncheng Li, Zehao Xue, Yingying Wang, Jianfei Cai, and
  Junsong Yuan.
\newblock {3D} hand shape and pose estimation from a single {RGB} image.
\newblock In {\em CVPR}, 2019.

\bibitem{hampali2020honnotate}
Shreyas Hampali, Mahdi Rad, Markus Oberweger, and Vincent Lepetit.
\newblock {HOnnotate: A method for 3d annotation of hand and object poses}.
\newblock In {\em CVPR}, 2020.

\bibitem{hasson2020leveraging}
Yana Hasson, Bugra Tekin, Federica Bogo, Ivan Laptev, Marc Pollefeys, and
  Cordelia Schmid.
\newblock Leveraging photometric consistency over time for sparsely supervised
  hand-object reconstruction.
\newblock In {\em CVPR}, 2020.

\bibitem{hasson2019learning}
Yana Hasson, Gul Varol, Dimitrios Tzionas, Igor Kalevatykh, Michael~J Black,
  Ivan Laptev, and Cordelia Schmid.
\newblock Learning joint reconstruction of hands and manipulated objects.
\newblock In {\em CVPR}, 2019.

\bibitem{he2016deep}
Kaiming He, Xiangyu Zhang, Shaoqing Ren, and Jian Sun.
\newblock Deep residual learning for image recognition.
\newblock In {\em CVPR}, 2016.

\bibitem{huang2020hand}
Lin Huang, Jianchao Tan, Ji Liu, and Junsong Yuan.
\newblock Hand-transformer: non-autoregressive structured modeling for {3D}
  hand pose estimation.
\newblock In {\em ECCV}, 2020.

\bibitem{ke2018multi}
Lipeng Ke, Ming-Ching Chang, Honggang Qi, and Siwei Lyu.
\newblock Multi-scale structure-aware network for human pose estimation.
\newblock In {\em ECCV}, 2018.

\bibitem{kingma2014adam}
Diederik~P Kingma and Jimmy Ba.
\newblock Adam: A method for stochastic optimization.
\newblock In {\em ICLR}, 2015.

\bibitem{kulon2020weakly}
Dominik Kulon, Riza~Alp Guler, Iasonas Kokkinos, Michael~M Bronstein, and
  Stefanos Zafeiriou.
\newblock Weakly-supervised mesh-convolutional hand reconstruction in the wild.
\newblock In {\em CVPR}, 2020.

\bibitem{lin2021end}
Kevin Lin, Lijuan Wang, and Zicheng Liu.
\newblock End-to-end human pose and mesh reconstruction with transformers.
\newblock In {\em CVPR}, 2021.

\bibitem{lin2017feature}
Tsung-Yi Lin, Piotr Doll{\'a}r, Ross Girshick, Kaiming He, Bharath Hariharan,
  and Serge Belongie.
\newblock Feature pyramid networks for object detection.
\newblock In {\em CVPR}, 2017.

\bibitem{liu2021semi}
Shaowei Liu, Hanwen Jiang, Jiarui Xu, Sifei Liu, and Xiaolong Wang.
\newblock Semi-supervised {3D} hand-object poses estimation with interactions
  in time.
\newblock In {\em CVPR}, 2021.

\bibitem{moon2018v2v}
Gyeongsik Moon, Ju~Yong Chang, and Kyoung~Mu Lee.
\newblock {V2V-PoseNet}: Voxel-to-voxel prediction network for accurate {3D}
  hand and human pose estimation from a single depth map.
\newblock In {\em CVPR}, 2018.

\bibitem{moon2017holistic}
Gyeongsik Moon, Ju~Yong Chang, Yumin Suh, and Kyoung~Mu Lee.
\newblock Holistic planimetric prediction to local volumetric prediction for
  {3D} human pose estimation.
\newblock {\em arXiv preprint arXiv:1706.04758}, 2017.

\bibitem{moon2020i2l}
Gyeongsik Moon and Kyoung~Mu Lee.
\newblock {I2L-MeshNet: Image-to-lixel prediction network for accurate 3D human
  pose and mesh estimation from a single RGB image}.
\newblock In {\em ECCV}, 2020.

\bibitem{moon2020neuralannot}
Gyeongsik Moon and Kyoung~Mu Lee.
\newblock {NeuralAnnot}: Neural annotator for in-the-wild expressive {3D} human
  pose and mesh training sets.
\newblock {\em arXiv preprint arXiv:2011.11232}, 2020.

\bibitem{moon2020pose2pose}
Gyeongsik Moon and Kyoung~Mu Lee.
\newblock Pose2{P}ose: {3D} positional pose-guided {3D} rotational pose
  prediction for expressive {3D} human pose and mesh estimation.
\newblock {\em arXiv preprint arXiv:2011.11534}, 2020.

\bibitem{moon2020deephandmesh}
Gyeongsik Moon, Takaaki Shiratori, and Kyoung~Mu Lee.
\newblock {DeepHandMesh}: A weakly-supervised deep encoder-decoder framework
  for high-fidelity hand mesh modeling.
\newblock In {\em ECCV}, 2020.

\bibitem{moon2020interhand2}
Gyeongsik Moon, Shoou-I Yu, He Wen, Takaaki Shiratori, and Kyoung~Mu Lee.
\newblock {InterHand2.6M}: A dataset and baseline for {3D} interacting hand
  pose estimation from a single {RGB} image.
\newblock In {\em ECCV}, 2020.

\bibitem{newell2016stacked}
Alejandro Newell, Kaiyu Yang, and Jia Deng.
\newblock Stacked hourglass networks for human pose estimation.
\newblock In {\em ECCV}, 2016.

\bibitem{paszke2019pytorch}
Adam Paszke, Sam Gross, Francisco Massa, Adam Lerer, James Bradbury, Gregory
  Chanan, Trevor Killeen, Zeming Lin, Natalia Gimelshein, Luca Antiga, et~al.
\newblock Pytorch: An imperative style, high-performance deep learning library.
\newblock In {\em NeurIPS}, 2019.

\bibitem{romero2017embodied}
Javier Romero, Dimitrios Tzionas, and Michael~J Black.
\newblock Embodied hands: Modeling and capturing hands and bodies together.
\newblock In {\em ACM TOG}, 2017.

\bibitem{sarandi2018robust}
Istv{\'a}n S{\'a}r{\'a}ndi, Timm Linder, Kai~O Arras, and Bastian Leibe.
\newblock How robust is {3D} human pose estimation to occlusion?
\newblock {\em arXiv preprint arXiv:1808.09316}, 2018.

\bibitem{vaswani2017attention}
Ashish Vaswani, Noam Shazeer, Niki Parmar, Jakob Uszkoreit, Llion Jones,
  Aidan~N Gomez, {\L}ukasz Kaiser, and Illia Polosukhin.
\newblock Attention is all you need.
\newblock In {\em NeurIPS}, 2017.

\bibitem{xu20213d}
Xiangyu Xu and Chen~Change Loy.
\newblock {3D} human texture estimation from a single image with transformers.
\newblock In {\em ICCV}, 2021.

\bibitem{Yang_2021_ICCV}
Sen Yang, Zhibin Quan, Mu Nie, and Wankou Yang.
\newblock Trans{P}ose: Keypoint localization via transformer.
\newblock In {\em ICCV}, 2021.

\bibitem{Zheng_2021_ICCV}
Ce Zheng, Sijie Zhu, Matias Mendieta, Taojiannan Yang, Chen Chen, and Zhengming
  Ding.
\newblock {3D} human pose estimation with spatial and temporal transformers.
\newblock In {\em ICCV}, 2021.

\bibitem{zhou2020occlusion}
Lu Zhou, Yingying Chen, Yunze Gao, Jinqiao Wang, and Hanqing Lu.
\newblock Occlusion-aware siamese network for human pose estimation.
\newblock In {\em ECCV}, 2020.

\bibitem{zhu2019robust}
Meilu Zhu, Daming Shi, Mingjie Zheng, and Muhammad Sadiq.
\newblock Robust facial landmark detection via occlusion-adaptive deep
  networks.
\newblock In {\em CVPR}, 2019.

\end{thebibliography}


\begin{thebibliography}{10}\itemsep=-1pt

\bibitem{chao2021dexycb}
Yu-Wei Chao, Wei Yang, Yu Xiang, Pavlo Molchanov, Ankur Handa, Jonathan
  Tremblay, Yashraj~S Narang, Karl Van~Wyk, Umar Iqbal, Stan Birchfield, et~al.
\newblock {DexYCB}: A benchmark for capturing hand grasping of objects.
\newblock In {\em CVPR}, 2021.

\bibitem{choi2020pose2mesh}
Hongsuk Choi, Gyeongsik Moon, and Kyoung~Mu Lee.
\newblock {Pose2Mesh}: Graph convolutional network for {3D} human pose and mesh
  recovery from a {2D} human pose.
\newblock In {\em ECCV}, 2020.

\bibitem{garcia2018first}
Guillermo Garcia-Hernando, Shanxin Yuan, Seungryul Baek, and Tae-Kyun Kim.
\newblock First-person hand action benchmark with {RGB-D} videos and {3D} hand
  pose annotations.
\newblock In {\em CVPR}, 2018.

\bibitem{hampali2020honnotate}
Shreyas Hampali, Mahdi Rad, Markus Oberweger, and Vincent Lepetit.
\newblock {HOnnotate: A method for 3d annotation of hand and object poses}.
\newblock In {\em CVPR}, 2020.

\bibitem{hasson2020leveraging}
Yana Hasson, Bugra Tekin, Federica Bogo, Ivan Laptev, Marc Pollefeys, and
  Cordelia Schmid.
\newblock Leveraging photometric consistency over time for sparsely supervised
  hand-object reconstruction.
\newblock In {\em CVPR}, 2020.

\bibitem{hasson2019learning}
Yana Hasson, Gul Varol, Dimitrios Tzionas, Igor Kalevatykh, Michael~J Black,
  Ivan Laptev, and Cordelia Schmid.
\newblock Learning joint reconstruction of hands and manipulated objects.
\newblock In {\em CVPR}, 2019.

\bibitem{lin2021end}
Kevin Lin, Lijuan Wang, and Zicheng Liu.
\newblock End-to-end human pose and mesh reconstruction with transformers.
\newblock In {\em CVPR}, 2021.

\bibitem{liu2021semi}
Shaowei Liu, Hanwen Jiang, Jiarui Xu, Sifei Liu, and Xiaolong Wang.
\newblock Semi-supervised {3D} hand-object poses estimation with interactions
  in time.
\newblock In {\em CVPR}, 2021.

\bibitem{moon2020i2l}
Gyeongsik Moon and Kyoung~Mu Lee.
\newblock {I2L-MeshNet: Image-to-lixel prediction network for accurate 3D human
  pose and mesh estimation from a single RGB image}.
\newblock In {\em ECCV}, 2020.

\bibitem{spurr2020weakly}
Adrian Spurr, Umar Iqbal, Pavlo Molchanov, Otmar Hilliges, and Jan Kautz.
\newblock Weakly supervised {3D} hand pose estimation via biomechanical
  constraints.
\newblock In {\em ECCV}, 2020.

\bibitem{vaswani2017attention}
Ashish Vaswani, Noam Shazeer, Niki Parmar, Jakob Uszkoreit, Llion Jones,
  Aidan~N Gomez, {\L}ukasz Kaiser, and Illia Polosukhin.
\newblock Attention is all you need.
\newblock In {\em NeurIPS}, 2017.

\end{thebibliography}

\end{document}



\title{Supplementary Material for \\``HandOccNet: Occlusion-Robust 3D Hand Mesh Estimation Network"}

\renewcommand{\thetable}{S\arabic{table}}
\renewcommand{\thesection}{S\arabic{section}}
\renewcommand{\thefigure}{S\arabic{figure}}
\renewcommand{\theequation}{S\arabic{equation}}


\author{JoonKyu Park$^{1*}$ \quad Yeonguk Oh$^{1*}$ \quad Gyeongsik Moon$^{1*}$ \quad Hongsuk Choi$^{1}$ \quad Kyoung Mu Lee$^{1,2}$ 
\\$^{1}$Dept. of ECE \& ASRI, $^{2}$IPAI, Seoul National University, Korea\\
{\tt\small jkpark0825@snu.ac.kr,namepllet1@gmail.com,\{mks0601,redarknight,kyoungmu\}@snu.ac.kr}}

\maketitle

\blfootnote{$^*$ Authors contributed equally.}

In this supplementary material, we first describe the specifications of FIT and SET in Section~\ref{sup:model_spec}.
In Section~\ref{sup:mpjpe}, we show quantitative results of our HandOCCNet before procrustes alignment to further justify our model.
In Section~\ref{sup:dexycb}, we show comparisons on the Dex-YCB dataset, which not only presents severe hand-object occlusion but also contains larger data.
In Section~\ref{sup:comparisons}, we provide additional visual comparisons of hand mesh estimation with the proposed HandOccNet and other state-of-the-art methods.

\section{Specifications of FIT and SET}
\label{sup:model_spec}
FIT injects hand information into the correlated occlusion region, and SET refines~$\mathbf{F}_{\text{FIT}}$ by referencing the distant information from $\mathbf{F}_{\text{FIT}}$.
In this section, we cover the detailed specification of each Transformer-based module.

\subsection{FIT}
\algdef{SE}[SUBALG]{Indent}{EndIndent}{}{\algorithmicend\ }
\algtext*{Indent}
\algtext*{EndIndent}
\definecolor{commentcolor}{RGB}{110,154,155}   
\newcommand{\PyComment}[1]{\textcolor{commentcolor}{\# #1}}
\newcommand{\PyCode}[1]{\ttfamily\textcolor{orange}{\# #1}} 

\begin{algorithm}[t]

\caption{Pseudocode of FIT in a PyTorch-style}
\label{algo:fit}
	\begin{algorithmic}[1]
	    \State \textbf{class} FIT(nn.Module): 
	        \Indent
	        \State def \textunderscore \textunderscore init \textunderscore\textunderscore():
	            \Indent
	            \State Q \textunderscore soft = nn.Conv2d(256, 256, 1)
	            \State K \textunderscore soft = nn.Conv2d(256, 256, 1)
	            \State Q \textunderscore sig = nn.Conv2d(256, 256, 1)
	            \State K \textunderscore sig = nn.Conv2d(256, 256, 1)
	            \State V = nn.Conv2d(256, 256, 1)
	            \State $\eta$ : $\mathbb{R}^{256\times 32\times 32}$ $\xrightarrow[]{}$ $\mathbb{R}^{1024\times256}$  \PyComment{reshape}
	            \State $\psi$ : $\mathbb{R}^{1024\times256}$ $\xrightarrow[]{}$ $\mathbb{R}^{256\times 32\times 32}$ \PyComment{reshape}
	            \State $\text{softmax}$ = nn.Softmax(dim=-1)
	            \State $\text{pool}$ = nn.AvgPool1d(1024, dim=-1)
	            \State $\text{LN}$ = nn.LayerNorm(256, dim=-1)
	            \State $\text{MLP}$ = nn.Sequential(\\
	            {\color{white}aaaaaaaaaaaaaaaaaaaa} nn.Linear(256, 256*4),\\
	            {\color{white}aaaaaaaaaaaaaaaaaaaa} nn.Linear(256*4, 256))
	            \State $d_{k_{\text{soft}}}, d_{k_{\text{sig}}} = 256, 256$
	            \EndIndent
	        \EndIndent
	        
	        \Indent
	        \State def forward($\mathbf{F}_{\text{S}}$, $\mathbf{F}_{\text{P}}$): \PyComment{$\mathbf{F}_{\text{S}} \text{ and } \mathbf{F}_{\text{P}} \in \mathbb{R}^{256\times 32\times 32}$}
    	        \Indent
    	        \State \PyComment{get queries, keys, and value}
    	        \State $q_{\text{soft}}$ = $\eta$(Q \textunderscore soft($\mathbf{F}_{\text{S}}$))
    	        \State $q_{\text{sig}}$ = $\eta$(Q \textunderscore sig($\mathbf{F}_{\text{S}}$))
    	        \State $k_{\text{sig}}$ = $\eta$(K \textunderscore sig($\mathbf{F}_{\text{P}}$))
    	        \State $k_{\text{sig}}$ = $\eta$(K \textunderscore sig($\mathbf{F}_{\text{P}}$))
    	        \State $v$ = $\eta$(V($\mathbf{F}_{\text{P}}$))
    	        
    	        \State \PyComment{softmax-based attention module}
    	        \State $\mathbf{C}_{\text{soft}}$ = 
    	        $\text{softmax}$(matmul$(q_{\text{soft}},{k_{\text{soft}}}^{T})/\sqrt{d_{k_{\text{soft}}}}$)
	            \State \PyComment{sigmoid-based attention module}
    
    	       \State $\mathbf{C}_{\text{sig}}$ = $\text{sigmoid}$(pool(matmul$(q_{\text{sig}},{k_{\text{sig}}}^{T})/\sqrt{d_{k_{\text{sig}}}}$))
    	       \State $\mathbf{C}$ = elemmul($\mathbf{C}_{\text{soft}}, \mathbf{C}_{\text{sig}}$)
    	       
                \State \PyComment{get residual feature~$\mathbf{R}_{\text{FIT}}$}
                \State $\mathbf{R}_{\text{FIT}}$ = matmul($\mathbf{C},v$)
                \State \PyComment{feed-forward module}
                \State $\mathbf{F}_{\text{FIT}} = \mathbf{F}_{\text{P}} + \psi(\mathbf{R}_{\text{FIT}}) + \psi(\text{MLP}(\text{LN}(\mathbf{R}_{\text{FIT}})))$

	        
	    \State \textbf{return} $\mathbf{F}_{\text{FIT}}$
	        \EndIndent
	        \EndIndent
	\end{algorithmic}
\end{algorithm}

We show the inference process of FIT in Algorithm~\ref{algo:fit}.
From the secondary feature~$\mathbf{F}_{\text{S}}$ and primary feature~$\mathbf{F}_{\text{P}}$, the  query and key features, ~$q_{\text{soft}}$ and $k_{\text{soft}}$ are computed by $1 \times 1$ convolution, respectively.
The query and key are matrix multiplied to produce an attention map, $\mathbf{C}_{\text{soft}}$.
To attenuate undesirable high attention scores from low matrix multiplication output in $\mathbf{C}_{\text{soft}}$, we compute an additional attention map~$\mathbf{C}_{\text{sig}}$ from matrix multiplication of additional query~$q_{\text{sig}}$ and key~$k_{\text{sig}}$.
All elements of $\mathbf{C}_{\text{sig}}$ range from 0 to 1.
$\mathbf{C}_{\text{soft}}$ and $\mathbf{C}_{\text{sig}}$ are multiplied
to produce a scaled attention map~$\mathbf{C}$, and matrix multiplied with a value feature from the primary feature~$\mathbf{F}_{\text{P}}$ to get the residual feature~$\mathbf{R}_{\text{FIT}}$.
Note that we do not use any residual connection to get $\mathbf{R}_{\text{FIT}}$.
The final output of FIT is obtained by feeding $\mathbf{R}_{\text{FIT}}$
into a feed-forward module with the residual connection between the module's input and output.
We also add a residual connection between its output and primary feature~~$\mathbf{F}_{\text{P}}$.

\subsection{SET}
\begin{algorithm}[t]
\caption{Pseudocode of SET in a PyTorch-style}
\label{algo:set}
	\begin{algorithmic}[1]
	    \State \textbf{class} SET(nn.Module): 
	        \Indent
	        \State def \textunderscore \textunderscore init \textunderscore\textunderscore()
	            \Indent
	            \State Q$'$ = nn.Conv2d(256, 256, 1)
	            \State K$'$ = nn.Conv2d(256, 256, 1)
	            \State V$'$ = nn.Conv2d(256, 256, 1)
	            \State $\eta$ : $\mathbb{R}^{256\times 32\times 32}$ $\xrightarrow[]{}$ $\mathbb{R}^{1024\times256}$ \PyComment{reshape}
	            \State $\psi$ : $\mathbb{R}^{1024\times256}$  $\xrightarrow[]{}$ $\mathbb{R}^{256\times 32\times 32}$ \PyComment{reshape}
	            \State $\text{softmax}$ = nn.Softmax(dim=-1)
	            \State $\text{LN}$ = nn.LayerNorm(256, dim=-1)
	            \State $\text{MLP}$ = nn.Sequential(\\
	            {\color{white}aaaaaaaaaaaaaaaaaaaa} nn.Linear(256, 256*4),\\
	            {\color{white}aaaaaaaaaaaaaaaaaaaa} nn.Linear(256*4, 256))
	            \State $d_{k'} = 256$
	            \EndIndent
	        \EndIndent
	        
	        \Indent
	        \State def forward($\mathbf{F}_{\text{FIT}}$): \PyComment{$\mathbf{F}_{\text{FIT}} \in \mathbb{R}^{256\times 32\times 32}$}
    	    
    	        \Indent
    	        \State \PyComment{get query, key, and value}
    	        \State $q'$ = $\eta$(Q$'$($\mathbf{F}_{\text{FIT}}$))
    	        \State $k'$ = $\eta$(K$'$($\mathbf{F}_{\text{FIT}}$))
    	        \State $v'$ = $\eta$(V$'$($\mathbf{F}_{\text{FIT}}$))
    	        
    	        \State \PyComment{softmax-based attention module}
    	        \State $\mathbf{C'}$ = 
    	        $\text{softmax}$(matmul$(q',{k'}^{T})/\sqrt{d_{k'}}$)
	            
	            \State \PyComment{get residual feature~$\mathbf{R}_{\text{SET}}$}
                \State $\mathbf{R}_{\text{SET}}$ = matmul($\mathbf{C'},v'$) + $q'$
                \State \PyComment{feed-forward module}
                \State $\mathbf{F}_{\text{SET}} = \psi(\mathbf{R}_{\text{SET}}) + \psi(\text{MLP}(\text{LN}(\mathbf{R}_{\text{SET}})))$

	        \State \textbf{return} $\mathbf{F}_{\text{SET}}$
	        \EndIndent
	        \EndIndent
	\end{algorithmic}
\end{algorithm}

In Algorithm~\ref{algo:set}, we show the inference process of our proposed SET.
Following the self-attention scheme in Transformer~\cite{vaswani2017attention}, we constrain the sum of attention values to be 1 by only adopting a softmax-based attention module.
We also follow the same pipeline of previous Transformers~\cite{vaswani2017attention} to get the residual feature~$\mathbf{R}_{\text{SET}}$ and the final output of SET, $\mathbf{F}_{\text{SET}}$.


\section{Evaluation: Before Procrustes Alignment}
\label{sup:mpjpe}
As results of hand mesh estimation before procrustes alignment are also significantly important in the literature, we further compare our HandOccNet to the state-of-the-art methods on the HO3D dataset in Table~\ref{sup_tab:mpjpe}.
As can be seen, our method still achieves better MPJPE before procrustes alignment.

\begin{table}[h]
    \vspace{-3mm}
    \centering
    \resizebox{1.0\linewidth}{!}{
    \begin{tabular}{c|ccccc}
        \toprule
        models & Pose2Mesh~\cite{choi2020pose2mesh} & Hasson~\etal~\etal~\cite{hasson2019learning} & I2L-MeshNet~\cite{moon2020i2l} & Liu~\etal~\cite{liu2021semi}] & \textbf{HandOccNet}\\
        \midrule
        MPJPE & 33.2 & 55.2 & 26.8 & 30.0 & \textbf{24.9} \\
        \bottomrule
    \end{tabular}}
    \caption{MPJPE before procrustes alignment comparison with state-of-the-art method on HO-3D.}
    \label{sup_tab:mpjpe}
    \vspace{-4mm}
\end{table}

\section{Results on the larger dataset, Dex-YCB~\cite{chao2021dexycb}}
\label{sup:dexycb}
We further compare our model to \cite{liu2021semi}, the second highest performing model on HO3D and FPHA datasets, in Table~\ref{sup_tab:dexycb}.
Dex-YCB consists of 582K RGB-D frames over 1,000 sequences of 10 subjects grasping 20 different objects from 8 independent views.
Therefore, evaluation on the Dex-YCB dataset could further justify our model's robustness to the situation where the hand is severely occluded.

\begin{table}[ht]
    \vspace{-3mm}
    \centering
    \resizebox{1.0\linewidth}{!}{
    \begin{tabular}{c|cccc}

        \toprule
        method &
        METRO~\cite{lin2021end} &
        Spurr~\etal~\cite{spurr2020weakly} & Liu~\etal~\cite{liu2021semi} & \textbf{HandOccNet} \\
        \midrule
        MPJPE & 15.24 & 17.34 & 15.28 &  \textbf{14.04} \\
        PA-MPJPE & 6.99 & 6.83 & 6.58 & \textbf{5.80}  \\
        \bottomrule
    \end{tabular}}
    \caption{Comparison with state-of-the-art methods on Dex-YCB dataset.}
    \label{sup_tab:dexycb}
    \vspace{-4mm}
\end{table}

\section{Qualitative comparisons}
\label{sup:comparisons}
\begin{figure*}[t]
\begin{center}
\includegraphics[width=\linewidth]{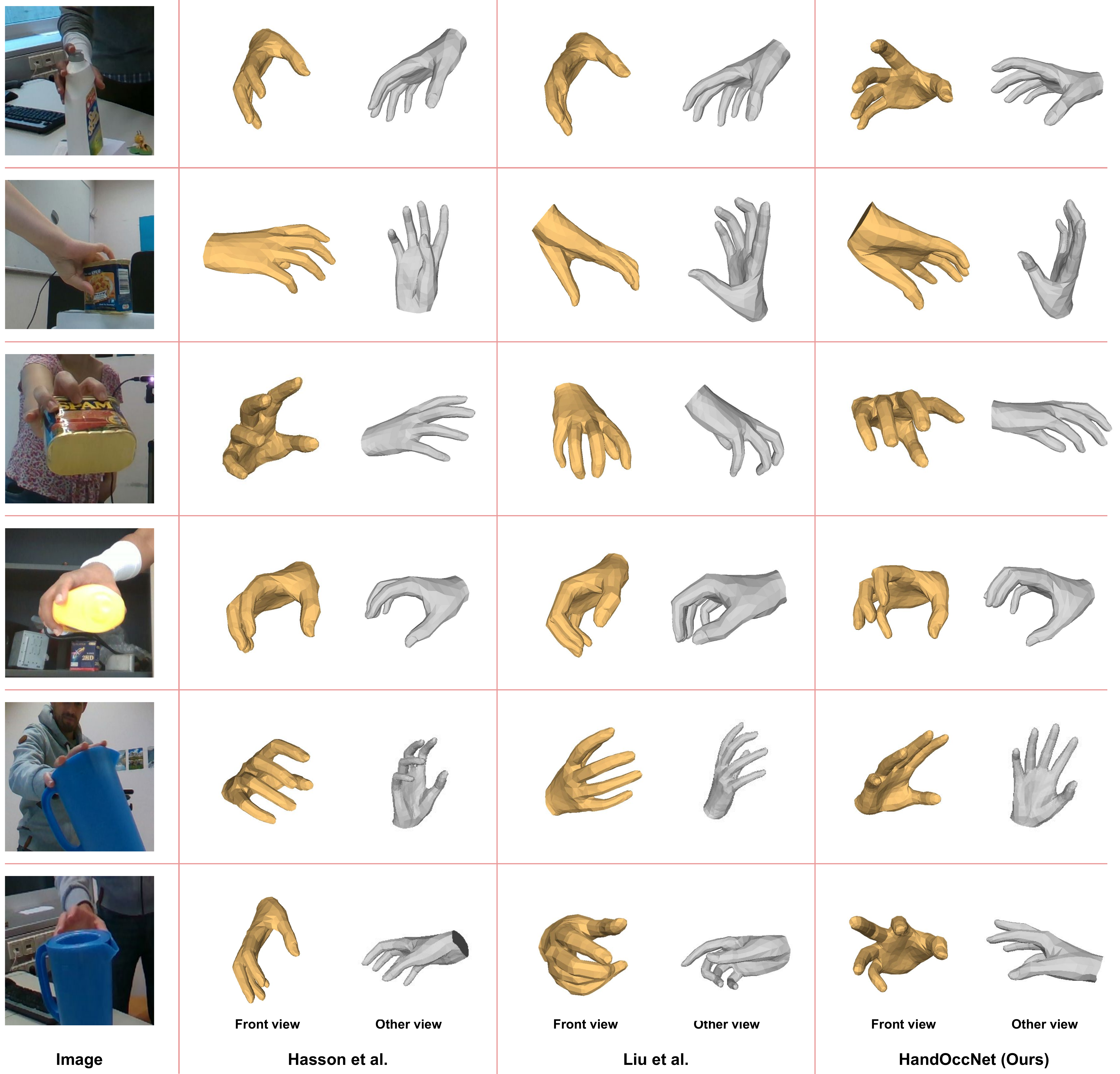}
\end{center}
\vspace*{-7mm}
\caption{
Qualitative comparisons of the proposed HandOccNet and state-of-the-art 3D hand mesh estimation methods~\cite{hasson2020leveraging, liu2021semi} on HO-3D~\cite{hampali2020honnotate}.
}
\vspace*{-4mm}
\label{sup_fig:HO3D}
\end{figure*}

\begin{figure*}[t]
\begin{center}
\includegraphics[width=\linewidth]{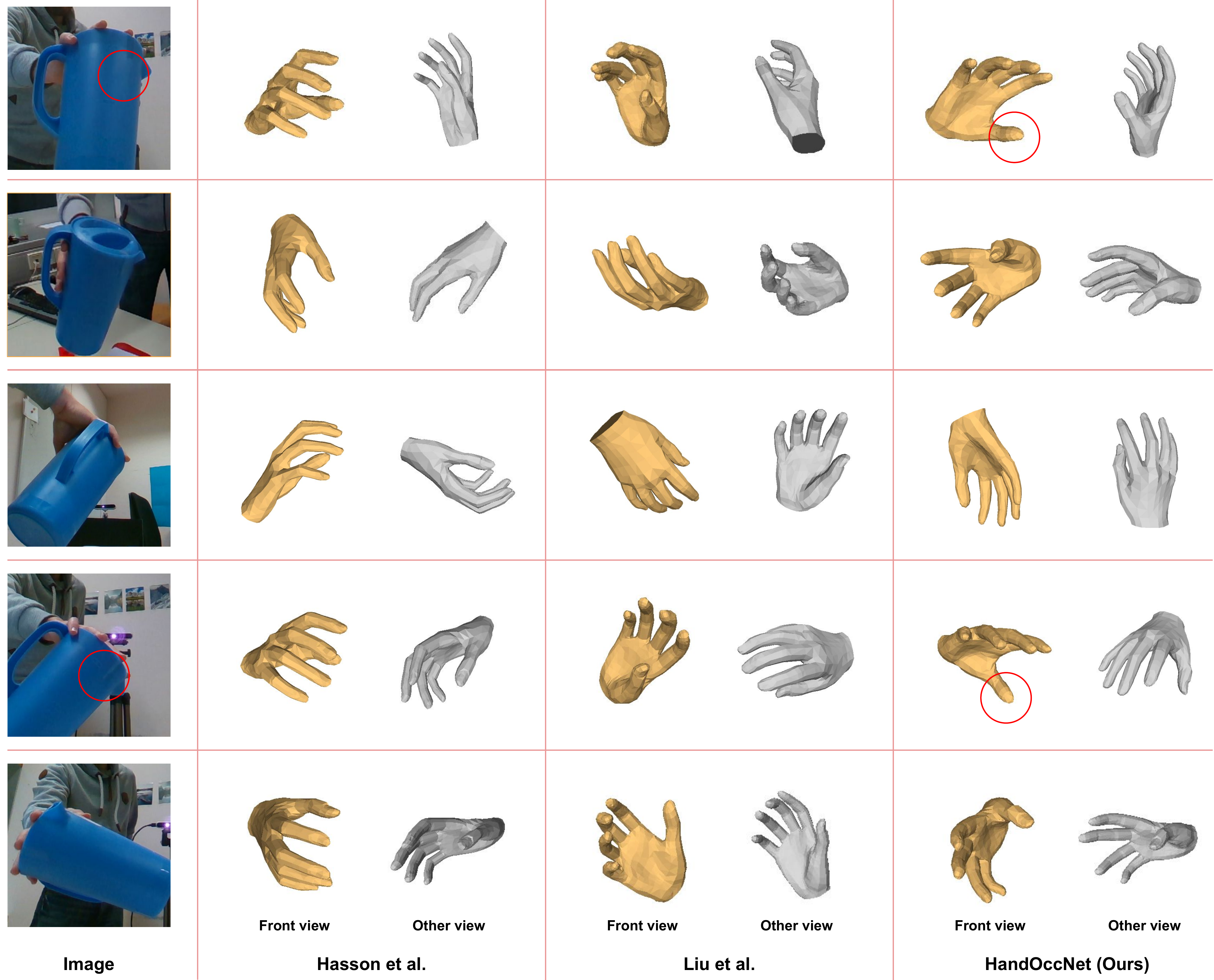}
\end{center}
\vspace*{-7mm}
\caption{
Qualitative comparisons of the proposed HandOccNet and state-of-the-art 3D hand mesh estimation methods~\cite{hasson2020leveraging, liu2021semi} on images of HO-3D~\cite{hampali2020honnotate} that contain severe occlusions.
}
\vspace*{-4mm}
\label{sup_fig:HO3D_severe}
\end{figure*}

\begin{figure*}[t]
\begin{center}
\includegraphics[width=0.9\linewidth]{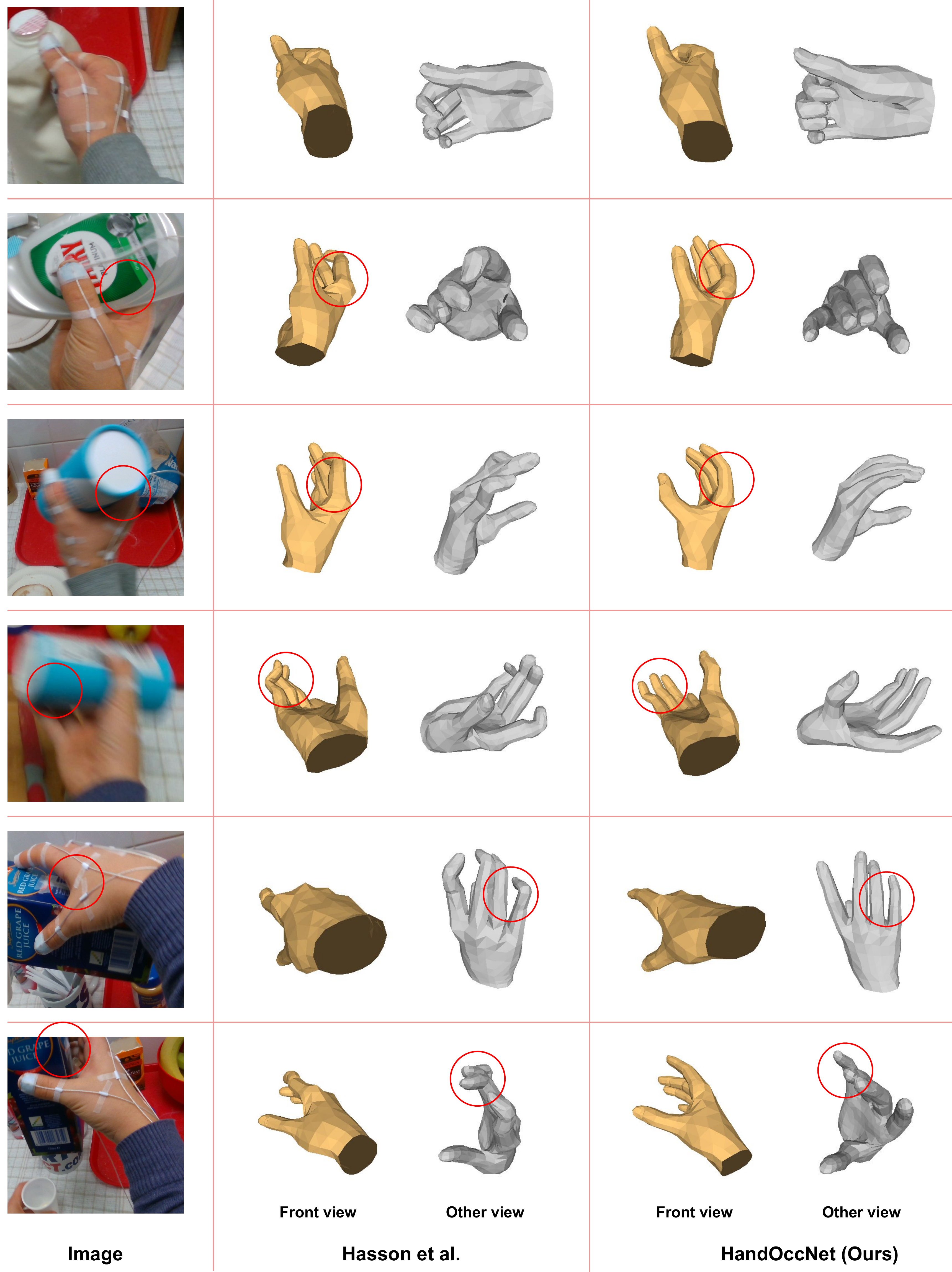}
\end{center}
\vspace*{-7mm}
\caption{
Qualitative comparisons of the proposed HandOccNet and state-of-the-art 3D hand mesh estimation methods~\cite{hasson2019learning} on FPHA~\cite{garcia2018first}.
}
\vspace*{-4mm}
\label{sup_fig:FPHA}
\end{figure*}

Figure~\ref{sup_fig:HO3D} shows more qualitative comparisons on HO-3D~\cite{hampali2020honnotate}.
Figure~\ref{sup_fig:HO3D_severe} further shows the qualitative comparisons on severely occluded images on HO-3D.
The figure shows that our HandOccNet can robustly estimate the 3D hand mesh when hands are severely occluded by objects.
This is due to our feature injection mechanism, which injects hand information into the occluded region and utilizes the injected information for the 3D hand mesh estimation.
For example, in the first row and fourth row in Figure~\ref{sup_fig:HO3D_severe}, thumbs are better reconstructed by injecting the relevant hand information into the occluded region.
Figure~\ref{sup_fig:FPHA} shows that our HandOccNet produces better results than Hasson~\etal~\cite{hasson2019learning} on FPHA~\cite{garcia2018first}.

\subsection*{License of the Used Assets}

\begin{compactitem}[$\bullet$]
    \item HO-3D dataset~\cite{hampali2020honnotate} is a publicly available dataset released under GNU GENERAL PUBLIC LICENSE v3.0.
    \item FPHA dataset~\cite{garcia2018first} is academically released dataset under Imperial College London. 
    \item \href{https://github.com/mks0601/I2L-MeshNet_RELEASE}{I2L-MeshNet~\cite{moon2020i2l} codes} are released for academic research only and it is free to researchers from educational or research institutes for non-commercial purposes.
    \item \href{https://github.com/hongsukchoi/Pose2Mesh_RELEASE}{Pose2Mesh~\cite{choi2020pose2mesh} codes} are released for academic research only and it is free to researchers from educational or research institutes for non-commercial purposes.
    \item \href{https://github.com/hassony2/handobjectconsist}{Hasson~\etal~\cite{hasson2020leveraging} codes} are released for academic research only and it is free to researchers from educational or research institutes for non-commercial purposes.
    \item \href{https://github.com/hassony2/obman_train}{Hasson~\etal~\cite{hasson2019learning} codes} are released for academic research only and it is free to researchers from educational or research institutes for non-commercial purposes.
    \item \href{https://github.com/stevenlsw/Semi-Hand-Object}{Liu~\etal~\cite{liu2021semi} codes} are released for academic research only and it is free to researchers from educational or research institutes for non-commercial purposes.
    \item \href{https://github.com/microsoft/MeshTransformer}{METRO~\cite{lin2021end} codes} are released under the MIT license.

\end{compactitem}

\clearpage
\clearpage
{\small
\bibliography{main}
}